\documentclass[journal]{IEEEtran}
\usepackage{amsmath,amsfonts}
\usepackage{algorithm}
\usepackage{array}
\usepackage{hyperref}
\usepackage{footnote}
\usepackage{tablefootnote}
\makesavenoteenv{tabular}
\makesavenoteenv{table}
\usepackage{keyval}
\usepackage{textcomp}
\usepackage{stfloats}
\usepackage{url}
\usepackage{svg}
\usepackage{verbatim}
\usepackage{bm}
\usepackage{bbold}
\usepackage{dsfont}
\usepackage{algorithm}
\usepackage[noend]{algpseudocode}
\usepackage{tikz}
\usepackage{relsize}								
\usepackage{framed} 
\usepackage[font={footnotesize}]{subcaption}
\usepackage[font={footnotesize}]{caption}
\usepackage{graphicx}
\usepackage{pgfplots}
\usepackage{multirow}
\usepackage{diagbox}
\usepackage{bigdelim, makecell, booktabs} 
\usepackage{enumitem} 
\pgfplotsset{compat=1.17}

\algnewcommand{\LeftComment}[1]{\Statex \(\triangleright\) #1}

\usepackage[noadjust]{cite}
\begin{document}
\bstctlcite{IEEEexample:BSTcontrol}

\title{Deep Statistical Solver for Distribution System State Estimation}

\author{
Benjamin~Habib, Elvin~Isufi,~\IEEEmembership{Member,~IEEE}, Ward~van~Breda, Arjen~Jongepier, Jochen~L.~Cremer,~\IEEEmembership{Member,~IEEE}

\thanks{Benjamin~Habib is with the Department of Electrical Sustainable Energy, TU Delft, Mekelweg 5, 2628 CD Delft, Netherlands. Email: \href{mailto:benjaminhabib@lilo.org}{benjaminhabib@lilo.org} }

\thanks{Elvin~Isufi 
is with the Department of Intelligent Systems, TU Delft, Mekelweg 5, 2628 CD Delft, Netherlands. Email: \href{mailto:E.Isufi-1@tudelft.nl}{E.Isufi-1@tudelft.nl}}

\thanks{Ward~van~Breda and Arjen~Jongepier are with Stedin Group, Blaak 8, 3011 TA Rotterdam, Netherlands
.}

\thanks{Jochen~L.~Cremer 
is with the Department of Electrical Sustainable Energy, TU Delft, Mekelweg 5, 2628 CD Delft, Netherlands. Email: \href{mailto:J.L.Cremer@tudelft.nl}{J.L.Cremer@tudelft.nl}}

\thanks{This research is supported by the TU Delft AI Initiative.}}

\markboth{Accepted at IEEE Transactions on Power Systems, June 2023}%
{Shell \MakeLowercase{\textit{et al.}}: Bare Demo of IEEEtran.cls for IEEE Transactions on Magnetics Journals}



\maketitle

\begin{abstract}
\textcolor{black}{Implementing accurate Distribution System State Estimation (DSSE) faces several challenges, among which the lack of observability and the high density of the distribution system. While data-driven alternatives based on Machine Learning models could be a choice, they suffer in DSSE because of the lack of labeled data. In fact, measurements in the distribution system are often noisy, corrupted, and unavailable. To address these issues, we propose the Deep Statistical Solver for Distribution System State Estimation (DSS\textsuperscript{2}), a deep learning model based on graph neural networks (GNNs) that accounts for the network structure of the distribution system and the governing power flow equations of the problem. 
DSS\textsuperscript{2} is based on GNN and leverages hypergraphs to model the network as a graph into the deep-learning algorithm and to represent the heterogeneous components of the distribution systems. A weakly supervised learning approach is put forth to train the DSS\textsuperscript{2}: by enforcing the GNN output into the power flow equations, we force the DSS\textsuperscript{2} to respect the physics of the distribution system. This strategy enables learning from noisy measurements and alleviates the need for ideal labeled data.  
Extensive experiments with case studies on the IEEE 14-bus, 70-bus, and 179-bus networks showed the DSS\textsuperscript{2} outperforms the conventional Weighted Least Squares algorithm in accuracy, convergence, and computational time while being more robust to noisy, erroneous, and missing measurements. The DSS\textsuperscript{2} achieves a competing, yet lower, performance compared with the supervised models that rely on the unrealistic assumption of having all the true labels.}
\end{abstract}


\vspace{-1em}
\section{Introduction}

\IEEEPARstart{D}{istribution} systems are taking a more active role in the energy transition. These active distribution systems require more extensive monitoring and control, which is possible by developing Distribution System State Estimation (DSSE) \cite{Lou22}. Currently, state estimation (SE) is mostly only possible in the transmission systems, and several challenges exist to extending SE to distribution systems successfully. First, conventional SE algorithms for transmission systems are challenging to adopt to distribution systems as the assumptions differ. Additionally, the distribution grid lacks real-time measurements. Conventional algorithms assume full observability of the grid with redundant measurements\textcolor{black}{, which is impractical} 
\cite{Primadianto2017}. \textcolor{black}{To address the observability issue of distribution systems,}forecasted values based on historical data \textcolor{black}{called \textit{pseudomeasurements}\footnote{It is common in DSSE to define pseudomeasurements as values estimated from historical data, as opposed to defining them as unmeasured variables calculated from the measurement of neighbor variables.} are used to} compensate for the lack of measurements, but they are often inaccurate and can impact the SE accuracy \cite{Giu14}. Also, the Weighted Least Squares (WLS) method used for SE is time-consuming and sensitive to data noise for large distribution systems \cite{fot22}. Multi-area SE has been widely investigated to speed up the estimation process \cite{Kek13, Kor11}, and has been extended to distribution systems \cite{Pau22}. However, the convergence and sensitivity issues remain, and division into multiple areas brings in communication and time-synchronization challenges. 

Different algorithms have been proposed to improve the robustness and convergence of SE, notably the branch-current WLS, the Least Absolute Value, and the Generalized Maximum Likelihood \cite{fot22,Dehghanpour2019}. Branch-current algorithms are more robust to parameter selection and uncertainty and are more suited for the weakly meshed and radial topologies in the distribution system \cite{Pau13}. Although, these algorithms suffer from the lack of qualitative measurements in wide distribution systems and the uncertainty of distributed loads and generators. Kalman Filters aim to improve speed and estimation performance under low observability. They are linked to the Forecast-Aided SE concept, where model-based approaches use the previous states as extra information to enhance accuracy and speed \cite{fot22,Ahmad2018,Jin2021,Liu2020,Kumari2021}. However, Kalman Filter SE is limited by the assumptions of system linearization and the Gaussian distribution of the measurements, which reduce its accuracy and robustness. Indeed, power systems are highly nonlinear, and measurements can show a non-Gaussian distribution \cite{Peg17_2}.  


Data-driven techniques showed promising results in performing fast DSSE without the above-mentioned assumptions. Deep learning models showed remarkable results to fit data for the SE task \cite{Mestav2019,Tran2021,Zargar2020, Jongh2020}. This \textit{supervised learning} approach trains neural networks to fit \textit{labels}, which are the grid's state variables. These labels are usually provided from simulations, as getting them from the grid is often impossible. As such, these approaches suffer from the scarcity of real labelled data and supervised learning is only possible using simulation data. Therefore, the models can only fit simulators exclusively and not real systems. Even though some approaches try to improve this technique by introducing some inductive bias \cite{Kun22} and physic-awareness \cite{Zam20}, they all require extensive supervised learning using large labelled databases \cite{fot22}. 

Combining model-based and data-driven approaches is a promising research direction to overcome the limitations of the model-based techniques with data-driven tools \cite{Jin2021}. This approach is considered in \cite{Revach2021} to combine the efficiency of Kalman Filters with the robustness of supervised Deep Learning architectures. It showed interesting results in low-dimensional problems; however, it suffers from high-dimensional problems due to the need for labelled data and unstable training. In the field of 'hybrid' approaches, \cite{Zhang19} combines data with physics and develops a model-specific deep neural network (DNN) by unrolling a SE solver to enhance estimation performance and alleviate computation expenses. However, the model is trained with fully labeled data, and the physic-awareness of the approach is limited and does not include the structure of the system.

\textcolor{black}{To address the lack of labeled data for training data-driven models, the concept of \textit{weakly supervised learning} has been proposed \cite{zho17}. This is similar to supervised learning but it is used for tasks where data is only partially or inaccurately labeled, relying on physical information, mathematical tools, or model-based modules to enhance the training process. While unsupervised learning also trains models without labels, weakly supervised learning still relies on imperfect values defined as labels. Although weakly supervised learning is highly practical for tasks such as DSSE, it has not yet been investigated for this particular problem.}

\textcolor{black}{Topology and parameter estimation in distribution systems from limited measurements have also been explored in the literature, with promising results. In \cite{Lave19}, a model-based algorithm using linear regression is proposed for topology and parameter estimation from limited measurements. In \cite{LiHa21}, smart meter data is utilized to estimate the topology through an Ordinary Least Squares method. Meanwhile, \cite{Liao19} proposes a data-driven Lasso algorithm for topology estimation.}

Graph Neural Networks (GNNs) are a particular family of deep learning models that use the underlying network structure as an inductive bias \cite{Gam20a, Bat18} to tackle the curse of dimensionality and reduce the data demand. GNNs have also shown robustness to perturbations in the network topology \cite{Gam20b,Gao21, Ma21}, which makes them appealing data-driven alternatives for the DSSE task. GNNs are investigated for power system applications, where the electrical lines correspond to the graph's edges and the buses correspond to the graph's nodes \cite{Lia21}, and the data varies for the specific application. 

\textcolor{black}{GNNs have been investigated for their potential use in SE in power systems, as shown in \cite{Rin21, Kun22, Kun22dist, Wu22}. The models in \cite{Rin21, Kun22} demonstrate that GNNs can accurately perform SE while being robust against noise and missing data. Moreover, \cite{Kun22dist} shows that GNNs can provide fast and robust SE, and any inaccuracies in the data would only impact local estimation. Additionally, \cite{Wu22} highlights that GNN models can handle fast sampling measurements, thereby improving SE in power systems. However, the heterogeneity of components in power systems cannot be accurately modeled using simple graphs, and thus, GNNs have limited expressivity in the graph model for power system applications.}

\textcolor{black}{Despite the increasing development of GNNs applications in power systems and growing research on deep learning for DSSE, the literature on GNN for DSSE is limited to parallel works \cite{Lin22, Mad23} In \cite{Lin22}, an electrical-model-guided GNN is used to perform DSSE and compared to conventional methods and other machine learning techniques. This approach demonstrated higher accuracy and robustness, indicating the potential of GNN-based approaches. In \cite{Mad23}, a GNN model is combined with matrix completion techniques to perform DSSE without the need for a detailed system model, highlighting the robustness of GNN approaches to model inaccuracies. While these approaches show promising results, they rely on labeled data for training, which is impractical due to the limited observability of the system's state.}

In this paper, we propose the \textit{Deep Statistical Solver for Distribution System State Estimation (DSS\textsuperscript{2})}, a GNN model based on the Deep Statistical Solver architecture \cite{Donon2022} specialized for optimization tasks on power systems. The model is trained in \textit{weakly supervision} manner to tackle the issues of data scarcity and inaccurate labeling. The success of such weak supervision is conveyed by considering \textcolor{black}{physical information of the network and }the physical laws of the power flow equations in the training loss function, rendering labels obsolete. Specific contributions include: 
\begin{enumerate}
    \item DSS\textsuperscript{2}, the Deep Statistical Solver model for accurate data-driven DSSE using a weakly supervised approach.
    \item adding physical constraints as penalization to the loss function, enhancing the model's performance.
    \item the innovative use of \textit{weakly supervision} in the context of data-driven DSSE, which leverages the power flow equations to restrict the model's search for the mapping function, hence reducing the data demand and improving robustness to inaccurate measurements.
\end{enumerate}

\textcolor{black}{Our goal with this proposition is to enhance the accuracy of grid estimation in situations where there are limited measurements and low-quality input values, such as pseudomeasurements. To achieve this, we propose using a deep learning model that learns solely from the grid's measurements and the physical information of the system.}

We validate the model using various case studies on the IEEE 14-bus, 70-bus, and 179-bus systems and compare it to the WLS algorithm baseline and other Deep Learning architectures. The proposed DSS\textsuperscript{2} is up to $15$ times faster, $4$ times more accurate, and more robust than the standard WLS algorithm. Our model also outperforms supervised learning approaches, being $10$ times more accurate in line-loading estimation while alleviating the need for labelled data. Interestingly, our approach is better for larger networks as the GNN learns in the neighborhood of buses, and the larger the power network, the more data to learn from. \textcolor{black}{The source code of this work is available on GitHub \cite{HabibGit}.}

This paper presents the methodology in Sections \ref{sec:DSS2} and \ref{sec:DSS}, introducing the Deep Statistical Solver model and extending its usage to DSS, respectively. Sec. \ref{sec:cs} are the case studies and compares the performances to the baseline algorithm and other data-driven models. Sec. \ref{sec:concl} concludes this work.

\section{Proposed approach for State Estimation}\label{sec:DSS2}
\subsection{Conventional problem formulation}

The state estimation problem aims at finding the state vector $\bm{x}$ based on a noisy measurement vector $\bm{z}$. Conventionally, we consider the voltage amplitude and angle at every grid bus as state variables, and $\bm{z}$ can include any measurement type: 
\vspace{-0.5em}
\small
\begin{subequations}
\begin{align}
\bm{x} &= \left[ V_0, V_1, \cdots, V_{n-1}, \varphi_0 = 0., \varphi_1, \cdots, \varphi_{n-1} \right] \, \in \, \mathbb{R}^{2n \times 1} \label{eq_state}\\
    \bm{z} &= \left[ z_0, z_1, \cdots, z_{m-1} \right] \, \in \mathbb{R}^{m \times 1}, \label{eq_meas}
\end{align}
\end{subequations}
\normalsize
where we consider $n$ buses and $m$ measurements. $V_i$ is the voltage amplitude at bus $i$, and $\varphi_i$ the voltage phase angle. We have $2n-1$ state variables, as $\varphi_0$ is set to $0$ by the slack bus convention. Linking the measurement vector $\bm{z}$ to the state vector $\bm{x}$, we define a measurement function $\bm{h}(\bm{x})$:
\vspace{-0.5em}
\small
\begin{equation}
    \bm{z} = \bm{h}(\bm{x}) + \bm{\varepsilon}, \label{link}
\end{equation}
\normalsize
where $\bm{\varepsilon} \in \mathbb{R}^{m \times 1}$ is the measurement noise vector, and $\bm{h}(\bm{x})$ are the power flow equations shown in Eq. \eqref{eq:meas_fun} \cite{Abur2004}. 

\vspace{-0.5em}
\scriptsize
\begin{equation} \label{eq:meas_fun}
\bm{h}(\bm{x}) = 
\begin{cases}
\begin{aligned}
V_i &= V_i \\

\varphi_i &= \varphi_i \\

P_{ij_\rightarrow} &= - V_i V_j \left[ \mathbb{R}(Y_{ij}) \cos(\Delta \varphi_{ij}) + \mathbb{I}(Y_{ij})\sin(\Delta \varphi_{ij}) \right] \\ &+ V_i^2 \left[\mathbb{R}(Y_{ij}) + \dfrac{\mathbb{R}(Y_{s_{ij}})}{2}\right] \\

P_{ij_\leftarrow} &= V_i V_j \left[ - \mathbb{R}(Y_{ij}) \cos(\Delta \varphi_{ij}) + \mathbb{I}(Y_{ij})\sin(\Delta \varphi_{ij}) \right] \\ &+ V_j^2 \left[\mathbb{R}(Y_{ij}) + \dfrac{\mathbb{R}(Y_{s_{ij}})}{2}\right] \\

Q_{ij_\rightarrow} &= V_i V_j \left[ - \mathbb{R}(Y_{ij}) \sin(\Delta \varphi_{ij}) + \mathbb{I}(Y_{ij})\cos(\Delta \varphi_{ij}) \right] \\ &- V_i^2 \left[\mathbb{I}(Y_{ij}) + \dfrac{\mathbb{I}(Y_{s_{ij}})}{2}\right] \\

Q_{ij_\leftarrow} &=  V_i V_j \left[ \mathbb{R}(Y_{ij}) \sin(\Delta \varphi_{ij}) + \mathbb{I}(Y_{ij})\cos(\Delta \varphi_{ij}) \right] \\ &- V_j^2 \left[\mathbb{I}(Y_{ij}) + \dfrac{\mathbb{I}(Y_{s_{ij}})}{2}\right] \\

I_{ij_\rightarrow} &= \left| \dfrac{P_{ij_\rightarrow} - jQ_{ij_\rightarrow}}{\sqrt{3}V_i e^{-j\varphi_i}}\right| =  \dfrac{\left| P_{ij_\rightarrow} - jQ_{ij_\rightarrow  }\right|}{\sqrt{3}V_i}\\

I_{ij_\leftarrow} &= \left| \dfrac{P_{ij_\leftarrow} - jQ_{ij_\leftarrow}}{\sqrt{3}V_j e^{-j\varphi_j}}\right| =  \dfrac{\left| P_{ij_\leftarrow} - jQ_{ij_\leftarrow  }\right|}{\sqrt{3}V_j}\\

P_{i} &= - \sum_{j \in \mathcal{N}(i)} P_{ij_\leftarrow} + P_{ij_\rightarrow} \\ 

Q_{i} &= - \sum_{j \in \mathcal{N}(i)} Q_{ij_\leftarrow} + Q_{ij_\rightarrow}
\end{aligned}
\end{cases}
\end{equation}
\normalsize
In this measurement function, Eq. \eqref{eq:meas_fun}, $\Delta \varphi_{ij} = \varphi_i - \varphi_j + \phi_{ij}$ is the voltage angle difference across the line that connects bus $i$ to bus $j$, $\phi_{ij}$ is the shift angle of the transformer if any, and $Y_{ij}$ and $Y_{s_{ij}}$ are respectively the line and shunt admittance of the line between bus $i$ and bus $j$. Measuring flows at bus $i$, we have $P_{ij_\rightarrow}$ and $Q_{ij_\rightarrow}$ as the active and reactive power flow from bus $i$ to bus $j$, and $P_{ij_\leftarrow}$ and $Q_{ij_\leftarrow}$ as the power flow from bus $j$ to bus $i$. Current flow $I_{ij}$ follows the same convention. Finally, we derive the active and reactive power injections at bus $i$, $P_i$, and $Q_i$ from the power flows. All these outputs are possible elements of $\bm{z}$, depending on the measurement infrastructure.

\textcolor{black}{The measurement function $\bm{h}(\bm{x})$ contains equations that connect the state variables $V_i$ and $\varphi_i$ to all types of measurements in the network. The first two lines correspond to identity functions that link the state variables to their direct measurements. Lines 3-6 consist of AC power flow equations used to derive power flows from the state variables. Lines 7-8 derive current flows from the previously derived power flows. Finally, lines 9-10 derive the nodes' power injection by ensuring exact power balance in each network node.}

\begin{figure}
    \centering    
    \begin{subfigure}[b]{0.42\textwidth}
    \includegraphics[width=\textwidth]{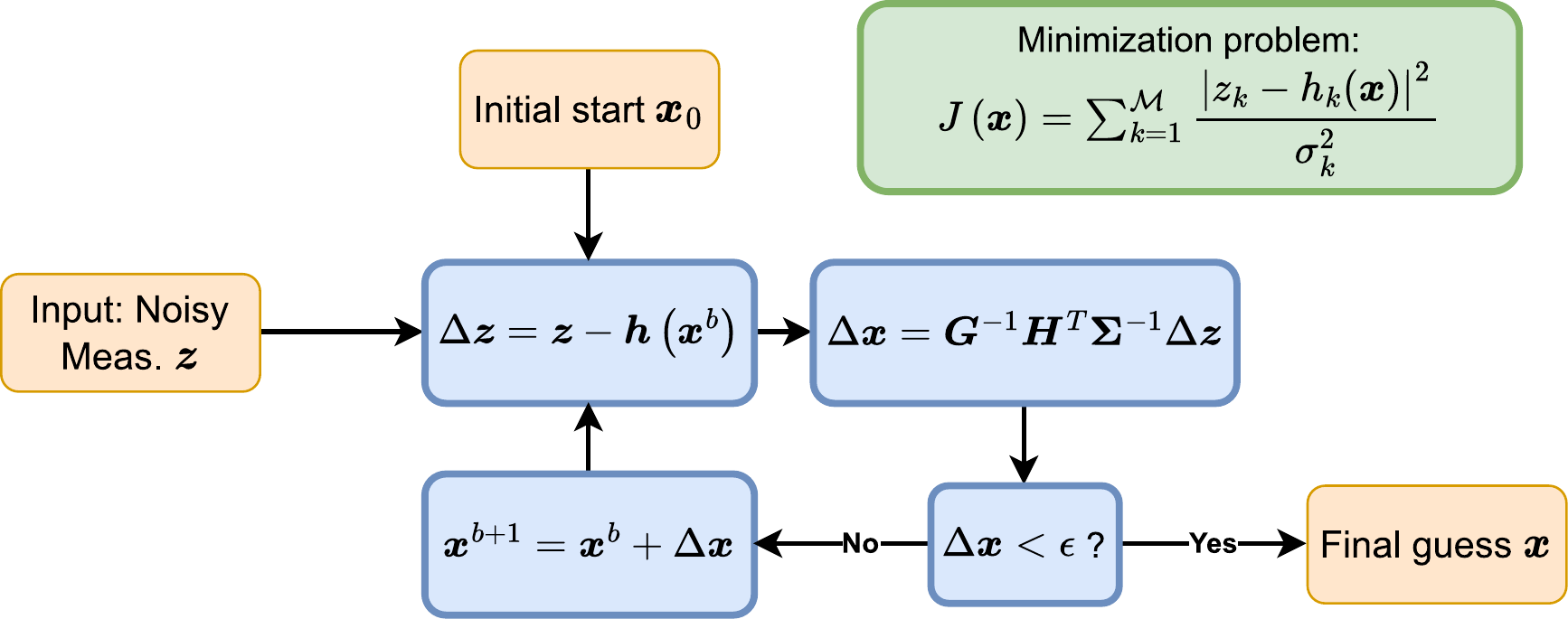}
    \caption{Weighted least squares approach}   
    \label{wls_approach}    
    \end{subfigure}    
    \begin{subfigure}[b]{0.42\textwidth}
    \includegraphics[width=\textwidth]{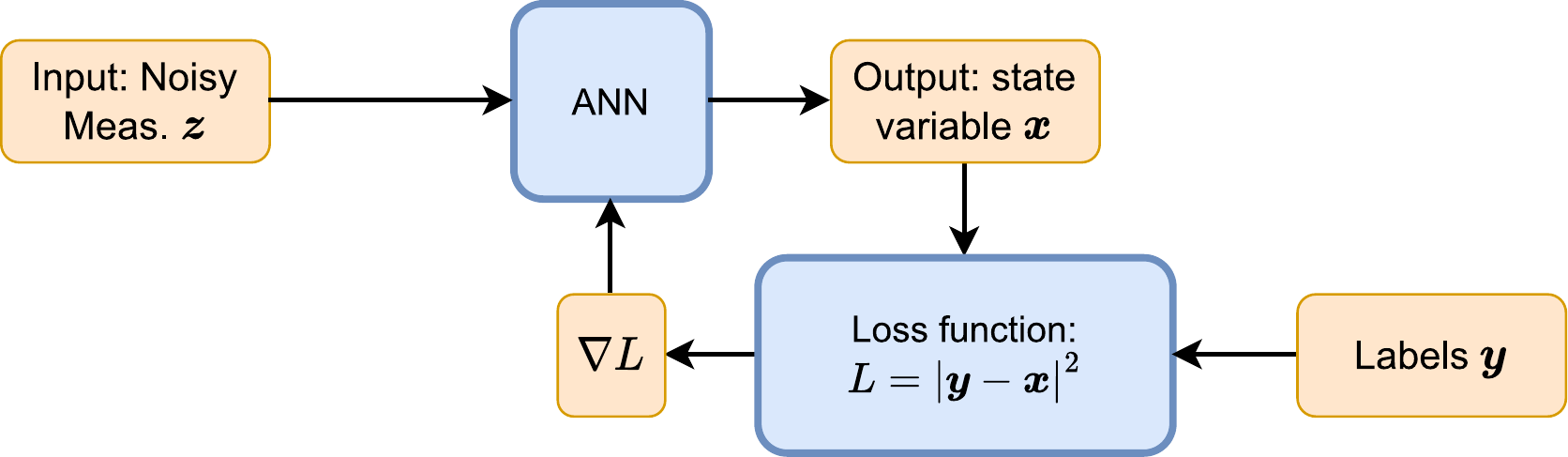}
    \caption{Supervised learning approach}
    \label{standard}
    \end{subfigure}    
    \begin{subfigure}[b]{0.42\textwidth}
    \includegraphics[width=\textwidth]{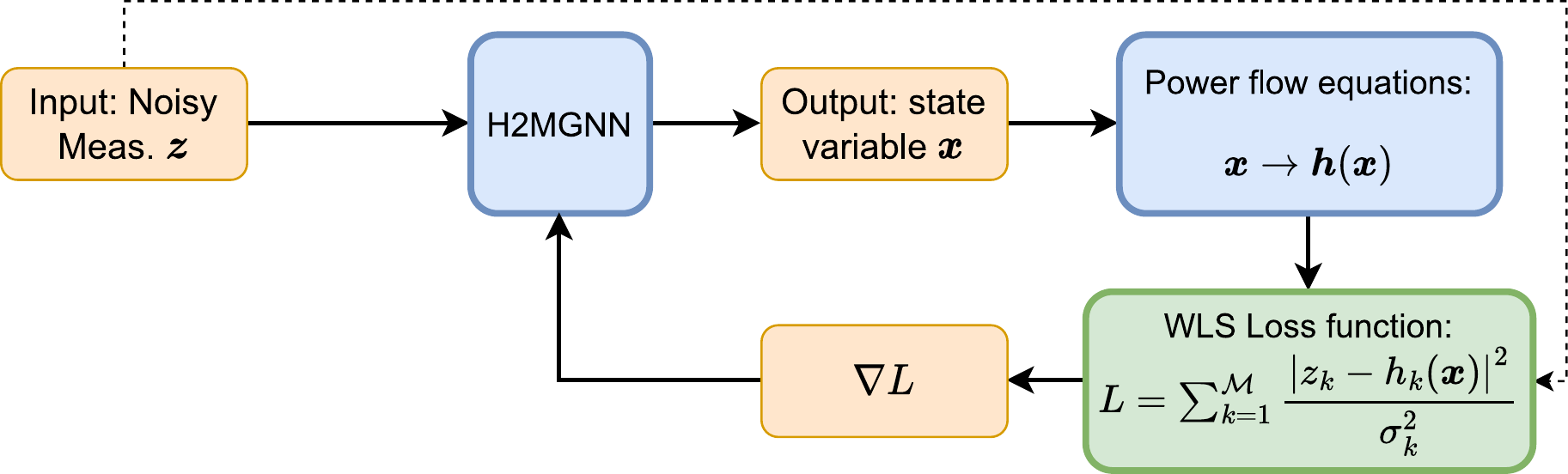}
    \caption{Proposed weakly supervised DSS$^2$ approach}
    \label{proposed_weak}
    \end{subfigure}    
    \caption{State estimation with (a) WLS with Newton-Raphson solver uses an initial guess $\bm{x}_0$ of the state vector $\bm{x}$ that iteratively updates $\bm{x}^b$ until the tolerance $\Delta \bm{x} > \epsilon$ or a maximum of iterations is reached, (b) supervised learning uses a label vector $\bm{y}$ to train an ANN to fit the output $\bm{x}$ to the input $\bm{z}$ and (c) the proposed weakly supervised approach considers the power flow equations $\bm{h}(\bm{x})$ (Eq. \ref{eq:meas_fun}) to get the \textit{estimated} measurements and fits the output $\bm{x}$ of the GNN model to the input $\bm{z}$ without labels. The target optimization is similar to WLS. \vspace{-2em}}
    \label{sup_unsup}
\end{figure}

The measurement function $\bm{h}(\bm{x})$ is nonlinear, and Eq. \eqref{link} includes the probabilistic noise vector $\bm{\varepsilon}$. In SE, we are interested in finding the inverse relation $\bm{h}^{-1}(\bm{z})$ to estimate the state vector $\bm{x}$ while compensating the error $\bm{\varepsilon}$. The conventional SE approach, shown in Fig. \ref{wls_approach}, uses the iterative Newton-Raphson algorithm to minimize a WLS objective function \cite{Abur2004}. This technique uses the redundancy of measurements to provide an accurate estimation. However, the approach requires at least the same number of measurements as state variables, meaning $ m \ge 2n-1$, and the system needs to be fully observable. Moreover, matching this requirement but failing to provide enough redundancy highly impacts the estimation accuracy. In practice, $ m \approx 4n $  achieves satisfying results, which is impractical for the distribution system \cite{Thurner2017}. The iterative process may even diverge in case of poor observability or high noise level in the measurements \cite{fot22}.

Another approach to approximate $\bm{h}^{-1}(\bm{z})$ is to train an Artificial Neural Network (ANN) to map this function. ANNs approximate functions using a series of nonlinear operations parameterized by their \textit{trainable weights} $\bm{\theta}$ \cite{Hor89}. These weights are estimated during the training phase to approximate the given relation. For the SE task, the model is trained to approximate the inverse relation $\bm{h}^{-1}(\bm{z})$, considering the measurement vector $\bm{z}$ as the input of the ANN and the state vector $\bm{x}$ as its output. We use the state estimation’s convention of $\bm{x}$ as output and $\bm{z}$ as input of the model.:
\vspace{-1em}

\begin{equation} \label{eq:mapping}
f_{\boldsymbol{\theta}}\left(\bm{z} \right) := \boldsymbol{h}^{-1}(\bm{z}) \rightarrow \bm{x}
\end{equation}

In a common supervised learning approach, the approach assigns a label vector $\bm{y}$ as the \textit{true value} of $\bm{x}$ for each measurement sample (one measurement vector $\bm{z}$). In the training process, shown in Fig. \ref{standard}, the model fits the data using available labels as reference. This approach, although quite efficient, is impractical for DSSE due to the lack of labelled data. Instead, our contribution combines Deep Learning and WLS optimization to propose a weakly supervised learning approach, alleviating the need for labels.

\vspace{-0.5em}
\subsection{Weakly supervised learning}

\textcolor{black}{To develop a weakly supervised learning approach for the DSSE task, we incorporate the power flow equations from Eq. \ref{eq:meas_fun} within the training phase, as shown in Fig. \ref{proposed_weak}. The H2MGNN algorithm takes the measurement vector $\bm{z}$ as an input, and provide an estimated state vector $\bm{x}$ as an output, which is then used to retrieve the estimated values $\bm{h}(\bm{x})$ of the network using the power flow equations (Eq. \eqref{eq:meas_fun}). The loss function of the model is then set to be the minimization objective of the WLS approach}:
\vspace{-0.5em}
\begin{equation} \label{eq:loss}
L(\bm{z}, \bm{x})=\sum_{k \in \mathcal{M}} \dfrac{\left| z_k - h_k(\boldsymbol{x}) \right|^2}{\sigma_k^2}
\end{equation}

\noindent with $\sigma_k$ the standard deviation of the  measurement $k$'s uncertainty assumed as Gaussian distribution, and $\mathcal{M}$ the measurements set. While SE aims at considering the actual noise $\varphi$ of the measurement vector, a `first guess´ of this value is assumed. $\sigma_k$. $|\mathcal{M}| = m$ is the number of measurements where $|\cdot|$ is the cardinality of a set. We assume uncorrelated measurements. 

\textcolor{black}{The detailed training procedure shown in Fig. \ref{proposed_weak} can be divided into four steps:
\begin{enumerate}
    \item An estimated state vector $\bf{x}$ is given by the H2MGNN algorithm from an input measurement vector $\bf{z}$.
    \item The power flow equations integrated into the measurement function $\bf{h}(\bf{x})$ are used to retrieve network's values from the estimated state $\bf{x}$.
    \item The \textit{estimated} values $\bf{h}(\bf{x})$ are compared to the actual measurement vector $\bf{z}$ in the Weighted Least Squared loss function. An estimation error is retrieved from each input measurement and weighted by the inverse of the measurement's variance.
    \item The sum of all estimation errors consists in the loss function, and we apply gradient descent to find the partial derivatives and tune the H2MGNN accordingly.
\end{enumerate}
}

\textcolor{black}{In this training process, the model is trained by minimizing the error between the measurements $\bm{z}$ and the estimated values $\bm{h}(\bm{x})$, and the uncertainty $\sigma^2$ of the measurements is used as weights to emphasize learning from the most accurate measurements. With this} loss function, we implement a weakly supervised learning approach where we use the input measurements $\bm{z}$ as \textit{noisy, imperfect labels} that the H2MGNN needs to fit through the power flow equations\textcolor{black}{, and no ground-truth labels are used for training}. Function \eqref{eq:meas_fun} is differentiable w.r.t the output state variables, and the gradient can be expressed using the measurement Jacobian matrix $ \bm{H}(\bm{x}) = \nabla \bm{h}(\bm{x})$.

\textcolor{black}{With this method, the target optimization of the training phase is exactly the conventional WLS minimization problem, allowing the model to learn an input-output mapping that represents this function. Our goal is to achieve a similar level of performance as the WLS approach while improving the model's numerical stability, computation time, robustness, and observability requirements.}

\subsection{Physical penalization terms in the loss function}
We propose aiding the WLS learning loss (\ref{eq:loss}) with different penalization terms to reduce the number of local minima and 'guide' the outputs towards physically-feasible solutions. \textcolor{black}{We assume our model does not need to estimate unstable states as protection schemes are faster and more reliable, so we guide the learning process to only estimate stable states in the output.} Considering stable networks, we add three terms to the loss:

\begin{itemize}
   \item Voltage level stability criteria: power systems ensure a voltage level between $V^{LB} = 95\%$ and $V^{UB} =105\%$ per unit to remain stable. Therefore, a two-sided penalization $[V-V^{UB}]_+ + [V^{LB}-V]_+$ is added to the loss function to enforce this criterion.\footnote{$[x]_+ = max(0,x)$}
    \item Phase angle stability criteria: large variations in phase angles are improbable in stable systems. For example, a phase angle difference of more than ${\Delta\varphi}^{UB} = 0.25$ rads between two neighbouring buses would characterize an unstable network. Therefore, we add a second two-sided penalization $[\Delta\varphi-{\Delta\varphi}^{UB}]_+ +[-{\Delta\varphi}^{UB}-\Delta\varphi]_+$ to the loss function to constrain this phase angle difference to ${\Delta\varphi}^{UB} = 0.25$.
    \item Line loading stability criteria: power systems regulators ensure the network's security by applying safety margins to line loading. To keep the model output within a physical range, we apply a third penalization $[l-l^{UB}]_+$ on the line loading when the prediction gives a loading higher than $l^{UB} =100\%$.
\end{itemize}

Adding these terms to the loss function, the equation used in the training process becomes:
\vspace{-1em}

\begin{equation}
\begin{aligned}
L(\bm{z}, \bm{x}) &=\sum_{k \in \mathcal{M}} \dfrac{\left| z_k - h_k(\boldsymbol{x}) \right|^2}{\sigma_k^2} + \lambda_0 [ \lambda_1 [V-V^{UB}]_+ \\ &+\lambda_1[V^{LB}-V]_+  +\lambda_2[\Delta\varphi-{\Delta\varphi}^{UB}]_+ \\ &+\lambda_2[-{\Delta\varphi}^{UB}-\Delta\varphi]_+ + \lambda_3[l-l^{UB}]_+ ] ,
\end{aligned}
\end{equation}

\noindent where $\lambda_0, \lambda_1, \lambda_2, \lambda_3$ are hyperparameters set to balance the effect of each mathematical term during training. These terms penalize the model output towards physically plausible boundaries and avoid diverging toward local minima that are well beyond the physical margins of the system. 

\begin{figure}
    \centering    
    \begin{subfigure}[b]{0.2\textwidth}
    \includegraphics[width=\textwidth]{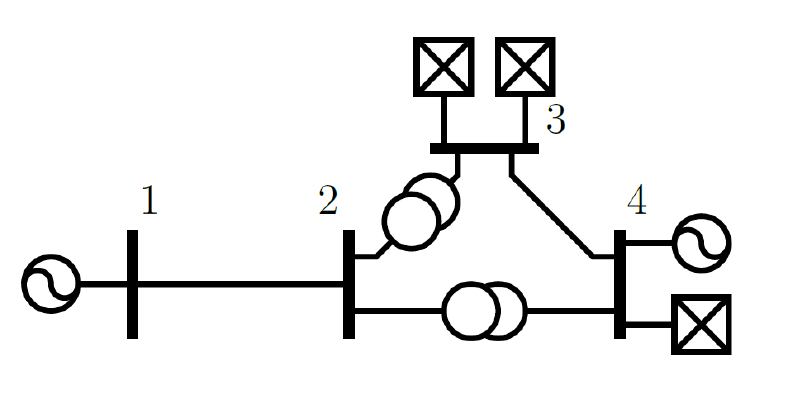}
    \caption{$4$-bus network}   
    \label{powernetwork}    
    \end{subfigure}    
    \begin{subfigure}[b]{0.2\textwidth}
    \includegraphics[width=\textwidth]{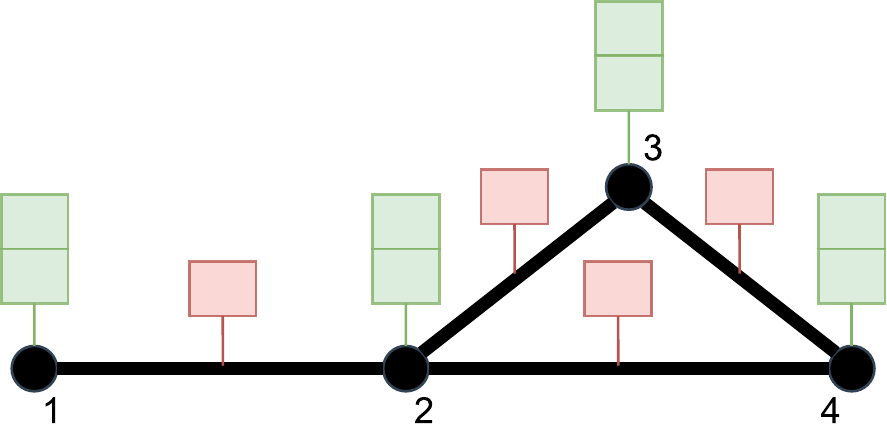}
    \caption{Standard graph model}
    \label{standard_graph}
    \end{subfigure}    
    \begin{subfigure}[b]{0.2\textwidth}
    \includegraphics[width=\textwidth]{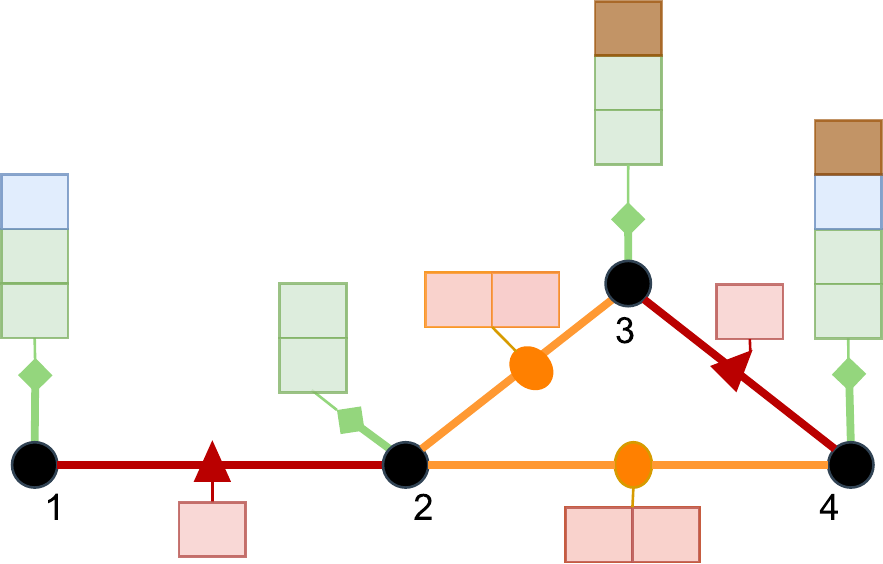}
    \caption{Hyper-Heterogeneous Multi Graph model (H2MG)}
    \label{h2mg}
    \end{subfigure}    
    \caption{Modelling the network (a) with two generators, three loads, two lines, and two transformers as (b) a standard graph and (c) an H2MG. The standard graph (b) has vertices (\protect\includegraphics[height=0.5em]{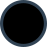}) and edges (\protect\includegraphics[height=0.1em]{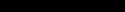}) with their features represented as boxes. 
    The H2MG (c) models the components, bus (\protect\includegraphics[height=0.5em]{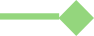}), line (\protect\includegraphics[height=0.5em]{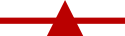}), and transformer (\protect\includegraphics[height=0.5em]{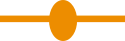}) as hyperedges connected to any number of connection ports with their features.\vspace{-1em}}
    \label{graph_vs_h2mg}
\end{figure}

\vspace{-0.5em}
\section{The Deep Statistical Solver model}\label{sec:DSS}

This section proposes the H2MG structure, the modelling of the heterogeneous components of the distribution grid, the Hyper-Heterogeneous Multi Graph Neural Network (H2MGNN) and how to learn the H2MGNN in weak supervision for DSS by applying Sec. \ref{sec:DSS2}.

\vspace{-0.5em}
\subsection{Hyper-Heterogeneous Multi Graph (H2MG)}
The H2MG uses hypergraphs to model power grids. Power grids are complex networks where different heterogeneous components are connected as shown in Fig. \ref{powernetwork}. Modelling power networks solely with vertices and edges, as done with standard graph models, Fig. \ref{standard_graph}, leads to information losses when merging grid components together into graph objects. More versatile modelling of such networks is possible using hypergraphs, Fig. \ref{h2mg}, where each component can be modelled as a specific hyperedge which can mitigate the loss of information. 


The H2MG formalism is defined by:

\begin{itemize}
    \item {Objects as hyperedges}: every object in the network is modelled as a hyperedge that can connect to any number of vertices. This is shown in Fig. \ref{graph_vs_h2mg} where each component is modelled separately as a hyperedge: we represent lines and transformers as hyperedges connected to two vertices, whereas buses are modelled as hyperedges connected to one vertex.
    \item {Vertices as ports}: vertices represent the interfaces between objects. In a hypergraph, vertices are \textit{connection points} between the components (the hyperedges). These connection points between components in a power system are the network's buses. Therefore, we model buses as both hyperedges as network components and vertices as network interfaces.
    \item {Hyper-Heterogeneous Multi Graph:} the collection of hyperedges connected through vertices forms a hypergraph, and we call this hypergraph \textit{heterogeneous} if it contains multiple classes of objects. 
    
    
    
\end{itemize}

Hyperedges carry features and outputs, while vertices, as connection ports, do not carry input-output information.

\vspace{-0.5em}
\subsection{H2MG Neural Network (H2MGNN)} \label{sec:h2mgnn}







The H2MGNN is a GNN architecture that works with H2MG models. It uses a recursive process to learn information from the hypergraph and related features. It is a recurrent and residual GNN architecture, with trainable mappings implemented as standard ANNs and trained through standard back-propagation. As presented in Algorithm \ref{alg:h2mgnn},  we consider four types of variables:

\begin{itemize}
    \item Vertex latent variables, considering a vertex set $\mathcal{V}$ corresponding to the \textit{interface} role of buses: $\bm{h}_{i}^{v}, \, \, \forall \, i \in \mathcal{V}$;
    \item hyperedge latent variables, considering $c \in \mathcal{C}$ as the objects' class, and $e \in \mathcal{E}^{c}$ as the objects' hyperedge: $\bm{h}_{e}^{c}$;
    \item hyperedge inputs $\bm{z}_{e}^{c}$;
    \item hyperedge outputs $\hat{\bm{x}}_{e}^{c}$.
\end{itemize}

In our setting, the hyperedge index $e$ refers to the object's connections: considering a vertex $i$ and its neighbouring vertex $j$, $e=i$ for a bus, and $e=ij$ for a line.

In the initialization of Algorithm \ref{alg:h2mgnn}, the hyperparameter $d$ sets the dimension of the latent variables. We initialize these latent variables with a flat start (zero values) and set predicted output variables to the initial values $\hat{\bm{x}}^c_{e,0}$ dependent on the task. For DSSE, a common initialization is $V_i =1$ p.u. and $\varphi_i = 0$rad. Then, the H2MGNN algorithm recursively updates these variables in the system with trainable mappings $\phi_{\theta}$. An iteration variable $t$ is defined to weigh each iteration in the update process and $T$ is the maximum number of iterations. At each iteration, latent variables are updated by an increment defined through the message-passing step similar to conventional GNN algorithms:
 \vspace{-1em}

 \begin{algorithm}
\caption{H2MGNN algorithm}\label{alg:h2mgnn}
\begin{algorithmic}[1]
\Procedure {$\boldsymbol{f}_{\theta}$}{$\bm{z}_{e}^{c}$}
\Statex
\LeftComment{Initialization}

\For {$i \in \mathcal{V}$}
    \State $\bm{h}_i^v \leftarrow \bm{0}^d$
\EndFor

\For {$c \in \mathcal{C}$}
    \For {$e \in \mathcal{E}^c$}
    \State $\bm{h}^c_{e} \leftarrow \bm{0}^d$
    \State $\hat{\bm{x}}^c_{e} \leftarrow \hat{\bm{x}}^c_{e,0}$
\EndFor
\EndFor
\Statex
\LeftComment{Latent interaction}
\State $t \leftarrow 0$

\While {$ t < T$}
    \For {$i \in \mathcal{V}$}
        \State $\bm{h}_{i}^{v} \leftarrow \bm{h}_{i}^{v}+\frac{1}{T} \times \sum \phi_{\theta}^{c, o}\left(\frac{t}{T}, \bm{h}_{i}^{v}, \bm{h}_{e}^{c}, \hat{\bm{x}}_{e}^{c}, z_{e}^{c}\right)$
    \EndFor
    
    \For {$c \in \mathcal{C}$}
    \For {$e \in \mathcal{E}^c$}
        \State $\bm{h}_{e}^{c} \leftarrow \bm{h}_{e}^{c}+\frac{1}{T} \times \phi_{\theta}^{c, h}\left(\frac{t}{T}, \bm{h}_{i}^{v}, \bm{h}_{e}^{c}, \hat{\bm{x}}_{e}^{c}, z_{e}^{c}\right)$
        \State $\hat{\bm{x}}_{e}^{c} \leftarrow \hat{\bm{x}}_{e}^{c}+ \, \, \frac{1}{T} \times \phi_{\theta}^{c, y}\left(\frac{t}{T}, \bm{h}_{i}^{v}, \bm{h}_{e}^{c}, \hat{\bm{x}}_{e}^{c}, z_{e}^{c}\right)$
    \EndFor
    \EndFor
    
    \State $ t \leftarrow t + 1$
\EndWhile
 \Statex      
\State \textbf{return} $\hat{\bm{x}}_{e}^{c}$
\EndProcedure
\end{algorithmic}
\end{algorithm}

\small
\begin{subequations}
\begin{align}
\Delta \bm{h}_{i}^{v}&=\sum_{(c, e, o) \in \mathcal{N}(i)} \phi_{\theta}^{c, o}\left(\frac{t}{T}, \bm{h}_{i}^{v}, \bm{h}_{e}^{c}, \hat{\bm{x}}_{e}^{c}, \bm{z}_{e}^{c}\right), \quad \forall i \in \mathcal{V} \label{GNN_algo_1}\\
\Delta \bm{h}_{e}^{c} &=\phi_{\theta}^{c, h}\left(\frac{t}{T}, \bm{h}_{i}^{v}, \bm{h}_{e}^{c}, \hat{\bm{x}}_{e}^{c}, \bm{z}_{e}^{c}\right) \label{GNN_algo_2} \\
\Delta \hat{\bm{x}}_{e}^{c} &=\phi_{\theta}^{c, y}\left(\frac{t}{T}, \bm{h}_{i}^{v}, \bm{h}_{e}^{c}, \hat{\bm{x}}_{e}^{c}, \bm{z}_{e}^{c}\right) \label{GNN_algo_3}
\end{align}
\label{GNN_algo}
\end{subequations}
\normalsize
with $\mathcal{N}(i)$ the set of hyperedges connected to vertex $i$, and $o$ the connection port of a hyperedge (if connected to multiple ports). The final output of the model is stored in the hyperedge outputs $\hat{\bm{x}}^c_e$.

\vspace{-1em}
\subsection{Proposed \texorpdfstring{DSS\textsuperscript{2}}{DSS2} implementation}
As presented in Fig. \ref{proposed_weak}, we use the H2MGNN model to estimate the state variables $\bm{x}$ from the measurements $\bm{z}$ through Alg. \ref{alg:h2mgnn} and train it through the weakly supervised approach with the WLS as target optimization. For the DSSE described in Sec. \ref{sec:DSS2}, the DSS\textsuperscript{2} model approximates the inverse of the measurement function Eq. \eqref{eq:meas_fun} as Eq. \eqref{eq:mapping}. \textcolor{black}{To simplify the model, we consider balanced systems and only model the positive sequence of the networks. We model buses, lines and transformers and integrate generators and loads as nodes' power injection, as commonly done for the DSSE task.}

The input features follow the WLS algorithm where, for each measurement, we consider the two, the measured value and its uncertainty. \textcolor{black}{Voltage angles are considered as  possible inputs of the model to allow the use of synchronized phasors, but are not required as most distribution systems do not carry such measurements.} We also add all other parameters as features needed to compute the measurement function Eq. (\ref{eq:meas_fun}) as \textit{topology parameters}. The features and parameters assigned to each class of components are listed in Table \ref{feat_tab}. The model's output is every bus's voltage amplitude and angle, as typical in SE. Finally, we add booleans to detail components: $\mathbb{1}_{z}$ defines buses with zero-injection (no consumption or generation), $\mathbb{1}_{s}$ defines slack buses, and $\mathbb{1}_{cl}$ defines closed lines. These booleans simplify the model and provide more information about the network to the DSS\textsuperscript{2} model. In other words, this simplification considers 'virtual measurements' to enforce zero power flow at buses without injection ($\mathbb{1}_{z}$), no power flow at the connected buses to an open line ($1-\mathbb{1}_{cl}$) and $V_s=1$ p.u. and $\phi_s=0$ rad at the slack bus $s$ where $s$ is the index of the vector $\mathbb{1}_{s}$ that equals $1$. \textcolor{black}{Since distribution grids typically have a limited number of measurements, we assume a low amount of measurements and incorporate pseudomeasurements to complete the observability of the system. These pseudomeasurements are based on historical demand data and are added as active and reactive power injections $P_i$ and $Q_i$ for buses where observability is lacking.}

\begin{table}
\caption{Features and topology parameters of the H2MGNN (modelled as an H2MG).}
\centering
\begin{tabular}{c|cc} 
\toprule
 & {Buses}                                                    & {Lines}                               \\ 
 \midrule
{\begin{tabular}[c]{@{}c@{}}Topology \\ param.\end{tabular}} & \begin{tabular}[c]{@{}c@{}}Bus port $i$ \\ Zero-inj. bool. $\mathbb{1}_{z}$    \\ Slack bool. $\mathbb{1}_{s}$    \end{tabular}                                                                                           & \begin{tabular}[c]{@{}c@{}} Line ports $(i,j)$ \\ Closed line bool. $\mathbb{1}_{cl}$    \\  Phase-shift $\phi_{ij}$    \end{tabular}  \\ 
\midrule
{\begin{tabular}[c]{@{}c@{}}Input \\ features\end{tabular}}      & \begin{tabular}[c]{@{}c@{}}Voltage magn.: $V_i$ - $\sigma_{V_i}$\\ Voltage angle: $\varphi_i$ - $\sigma_{\varphi_i}$\\ Active power inj.: $P_i$ - $\sigma_{P_i}$\\ Reactive power inj.: $Q_i$ - $\sigma_{Q_i}$\end{tabular} & \begin{tabular}[c]{@{}c@{}}Active PF: $P_{ij}$ - $\sigma_{P_{ij}}$\\ Reactive PF: $Q_{ij}$ - $\sigma_{Q_{ij}}$\\ Current magn.: $I_{ij}$ - $\sigma_{I_{ij}}$\\ Line admittance: $Y_{ij}$ \\ Shunt admittance: $Y_{s_{ij}}$\end{tabular} \\ 
\midrule
{\begin{tabular}[c]{@{}c@{}}Output \\ features\end{tabular}}     & \begin{tabular}[c]{@{}c@{}}Voltage magn.: $V_i$ \\Voltage angle: $\varphi_i$               \end{tabular}  & \\
\bottomrule
\end{tabular}
\label{feat_tab}
\end{table}

\vspace{-1em}
\section{Case studies}
\label{sec:cs}
Case studies have been undertaken to provide insights into the proposed approach and evidence of its efficacy. After stating the case studies settings and showing the efficiency of the proposed weakly-supervised learning approach, 
we analyse the performance of the DSS\textsuperscript{2} 
exploring the trade-off of providing labels and accuracy, subsequently, investigating the accuracy, convergence and computational speed for larger networks. 
Finally, we investigate the performance of the proposed approaches for different measurement noise, 
when the measurements are disturbed,
and when we have higher and lower load levels and renewable powers. 

\vspace{-0.5em}
\subsection{Test systems and setup}
\begin{figure}
    \centering    
    \begin{subfigure}[b]{0.25\textwidth}
    \includegraphics[width=\textwidth]{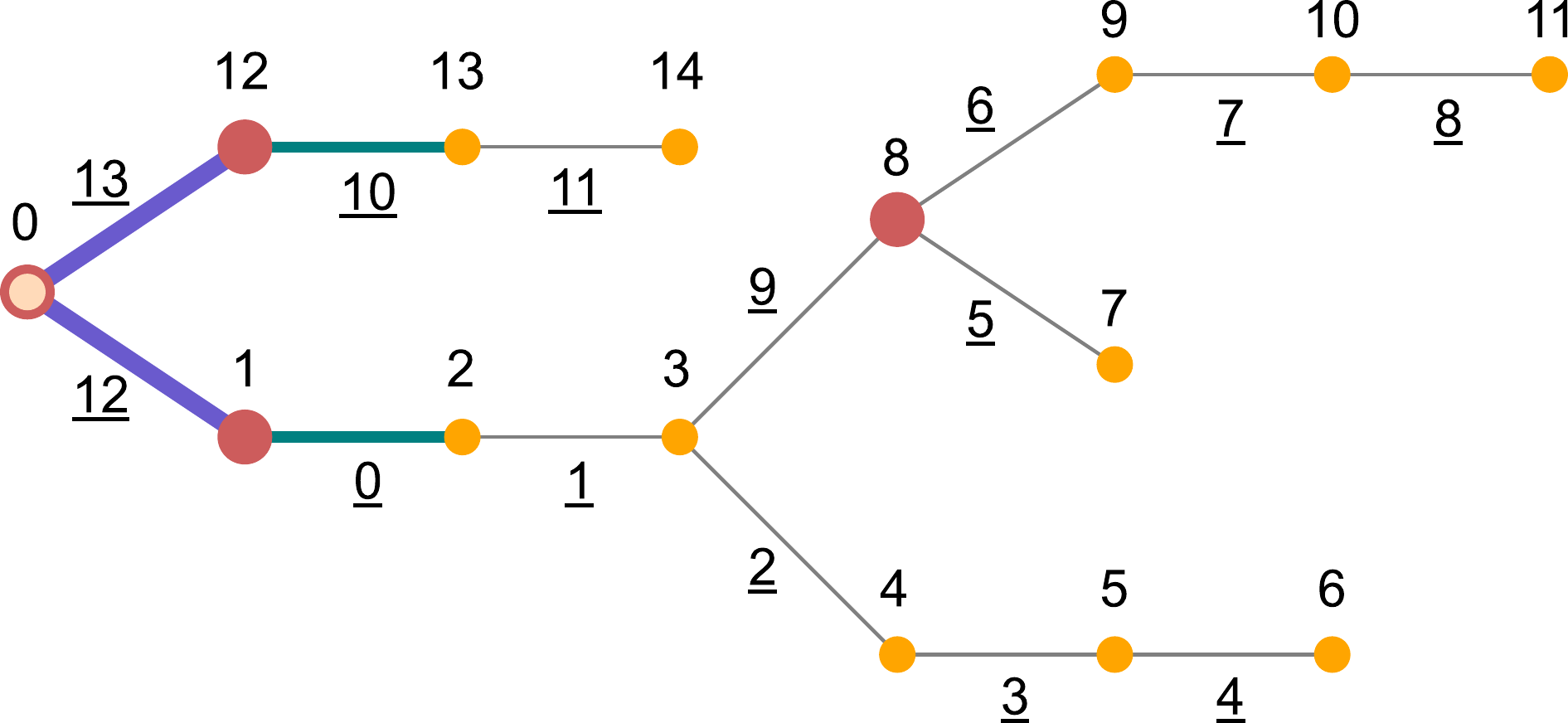}
    \caption{14-bus CIGRE MV grid from \cite{Barsali2014}}   
    \label{cigre14}    
    \end{subfigure}    
    \begin{subfigure}[b]{0.33\textwidth}
    \includegraphics[width=\textwidth]{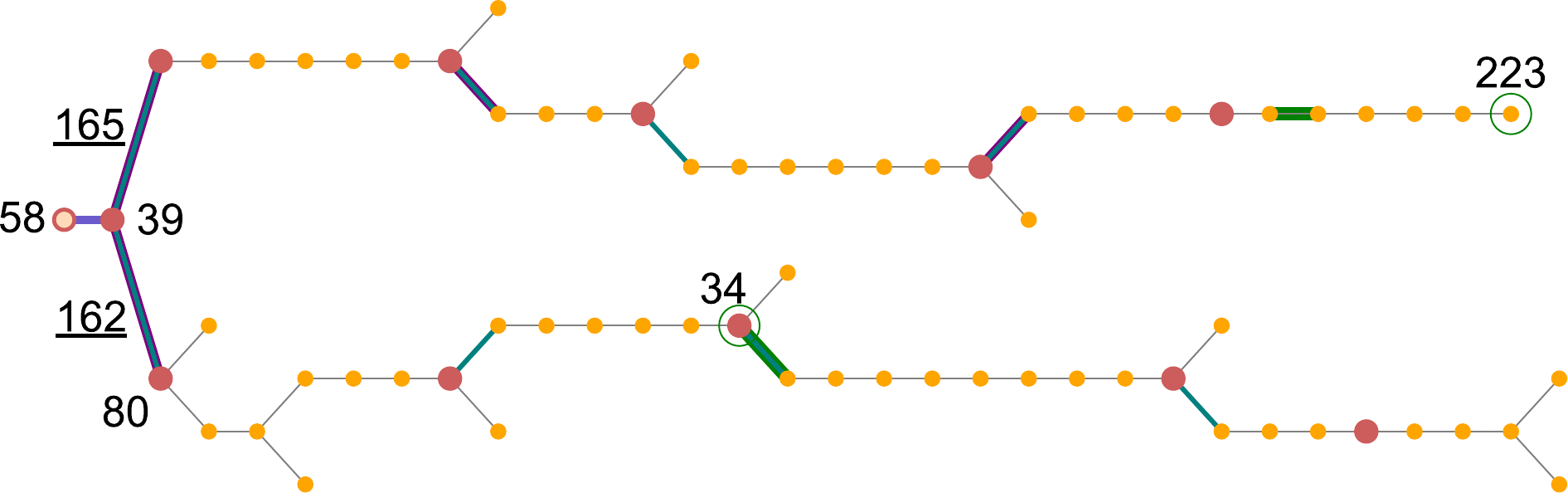}
    \caption{70-bus Oberrhein MV sub-grid from \cite{Thurner2017}
    }
    \label{ober70}
    \end{subfigure}    
    \begin{subfigure}[b]{0.33\textwidth}
    \includegraphics[width=\textwidth,height=14em]{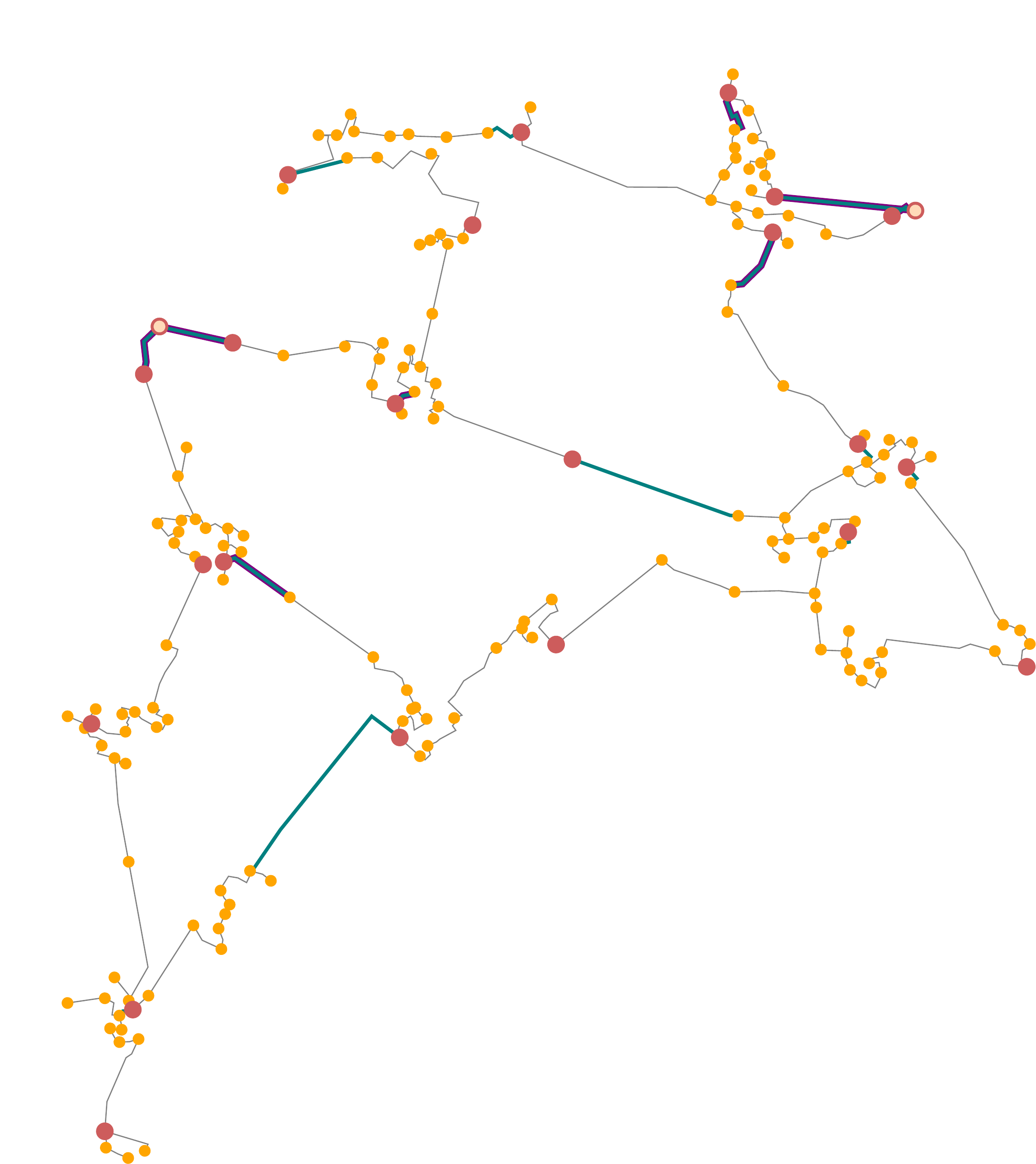}
    \caption{179-bus Oberrhein MV grid from \cite{Thurner2017}}
    \label{ober179}
    \end{subfigure}    
    \caption{Three test networks consisting out of trafos (\protect\includegraphics[height=0.5em]{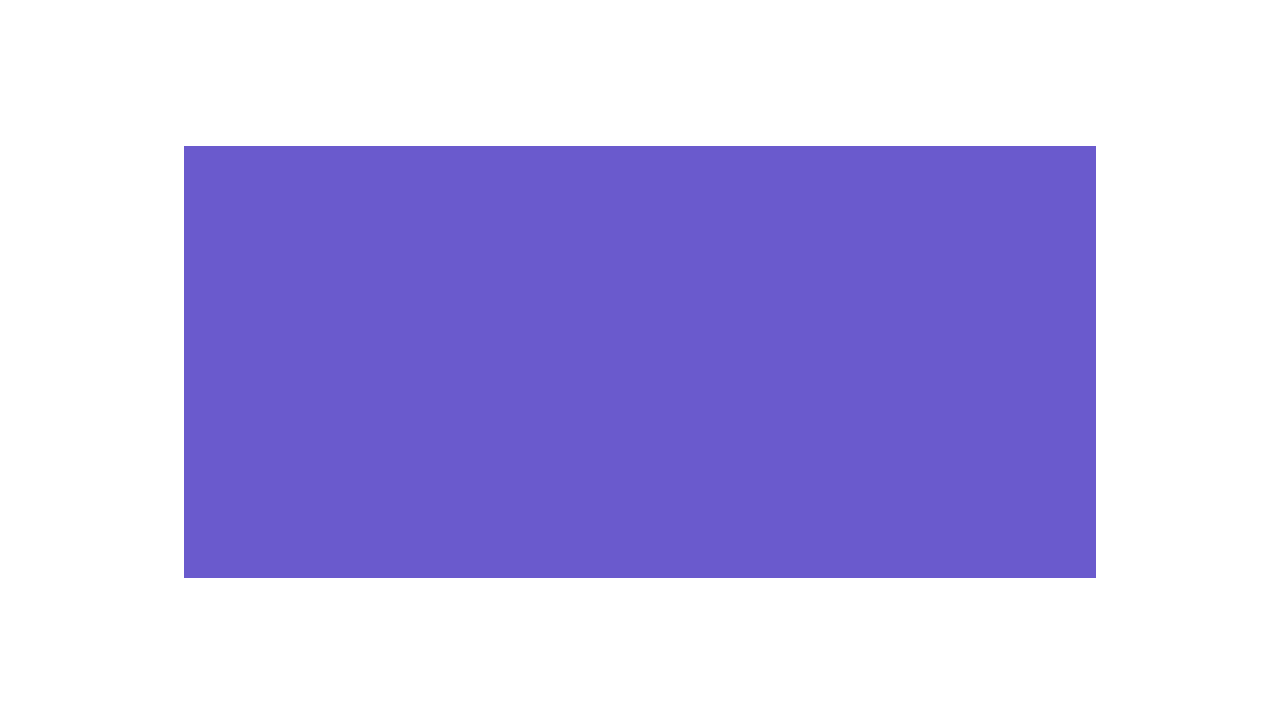}), lines (\protect\includegraphics[height=0.5em]{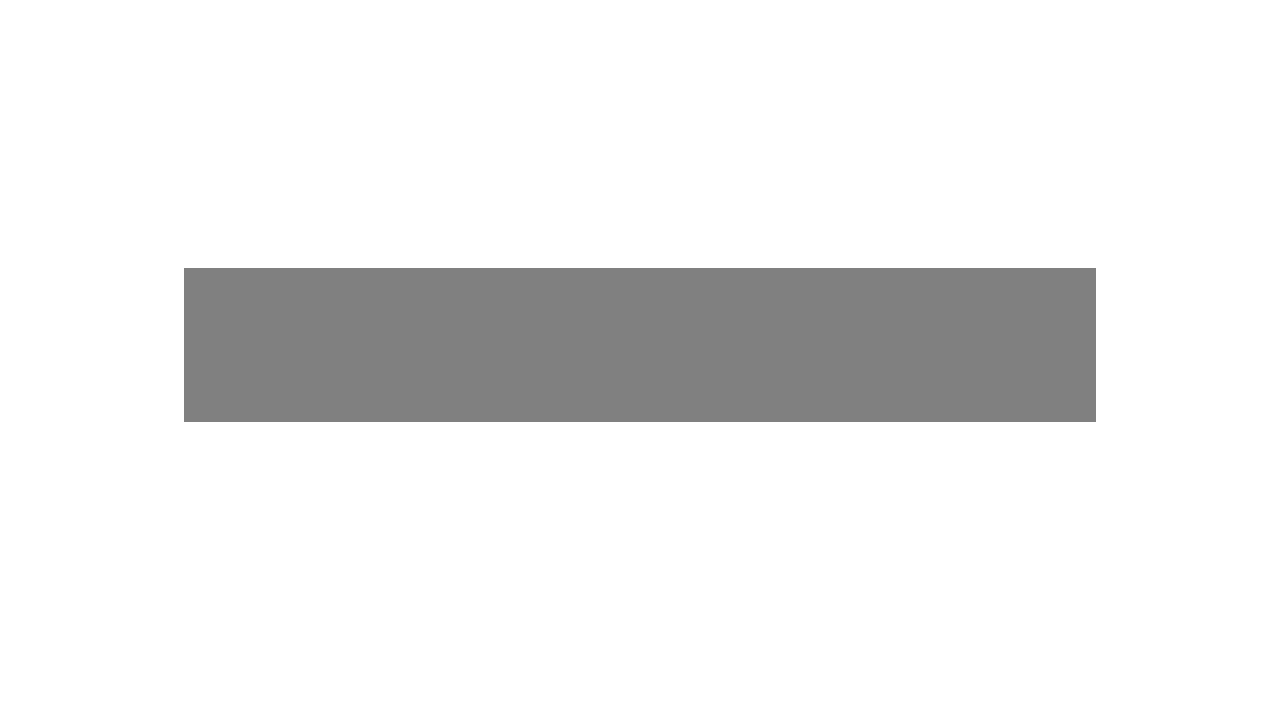}), MV/LV buses (\protect\includegraphics[height=0.5em]{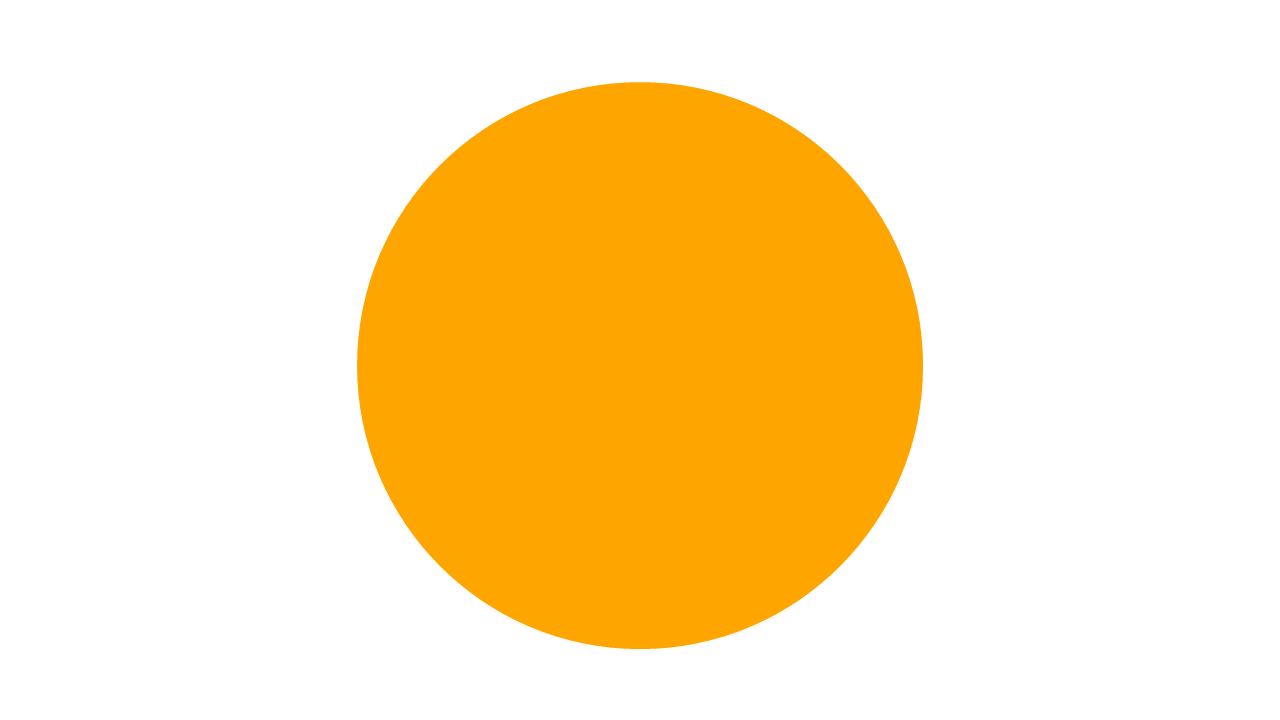}) and HV buses (\protect\includegraphics[height=0.5em]{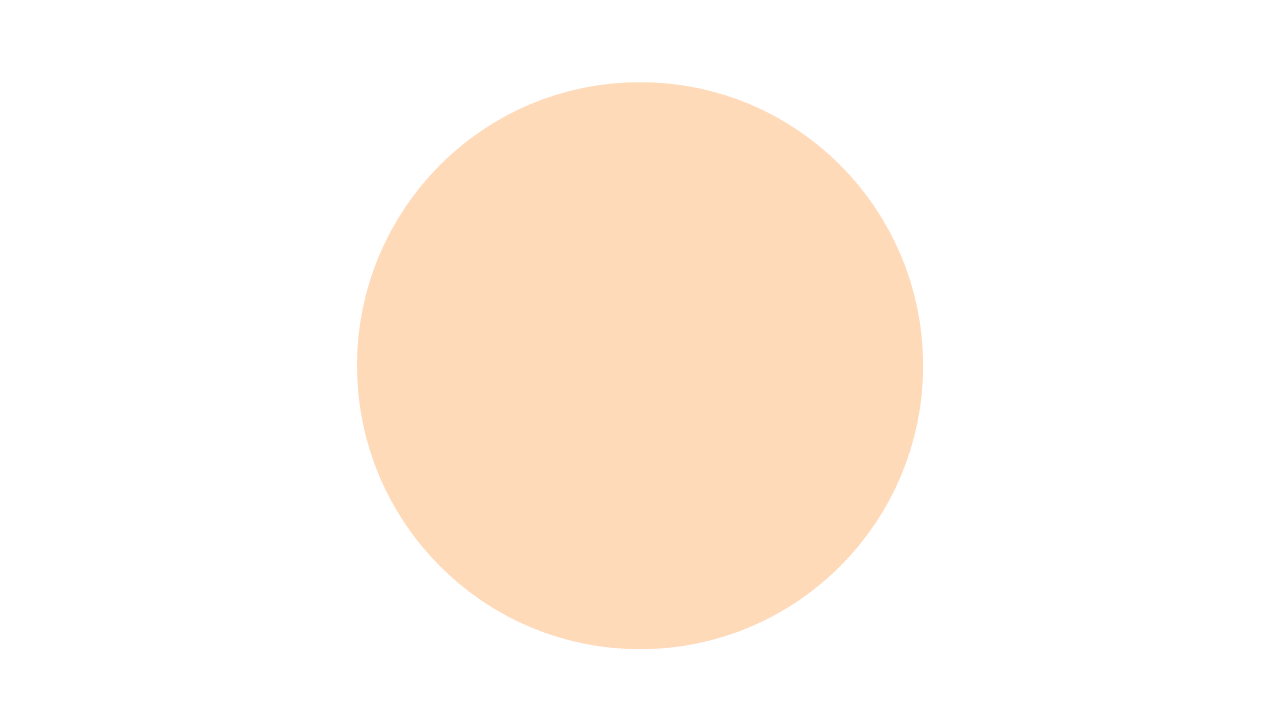}). The state can be estimated using the set of power flow measurements (\protect\includegraphics[height=0.5em]{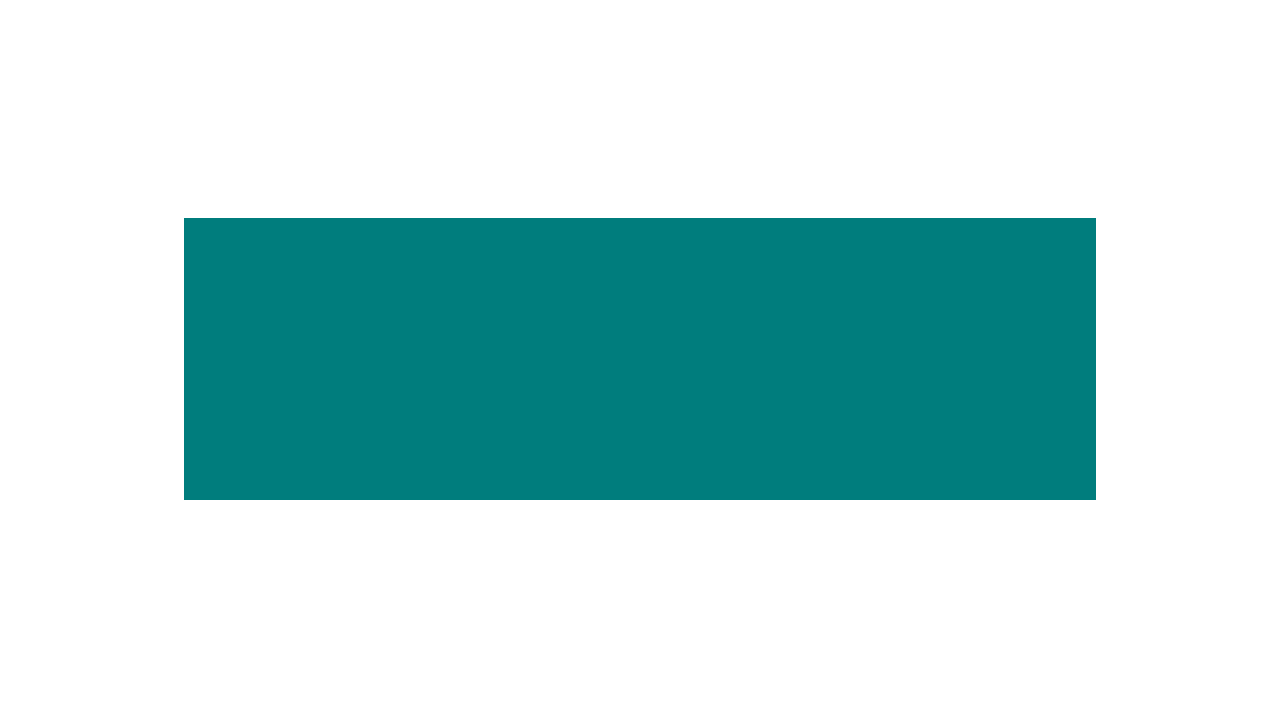}) and voltage measurements (\protect\includegraphics[height=0.5em]{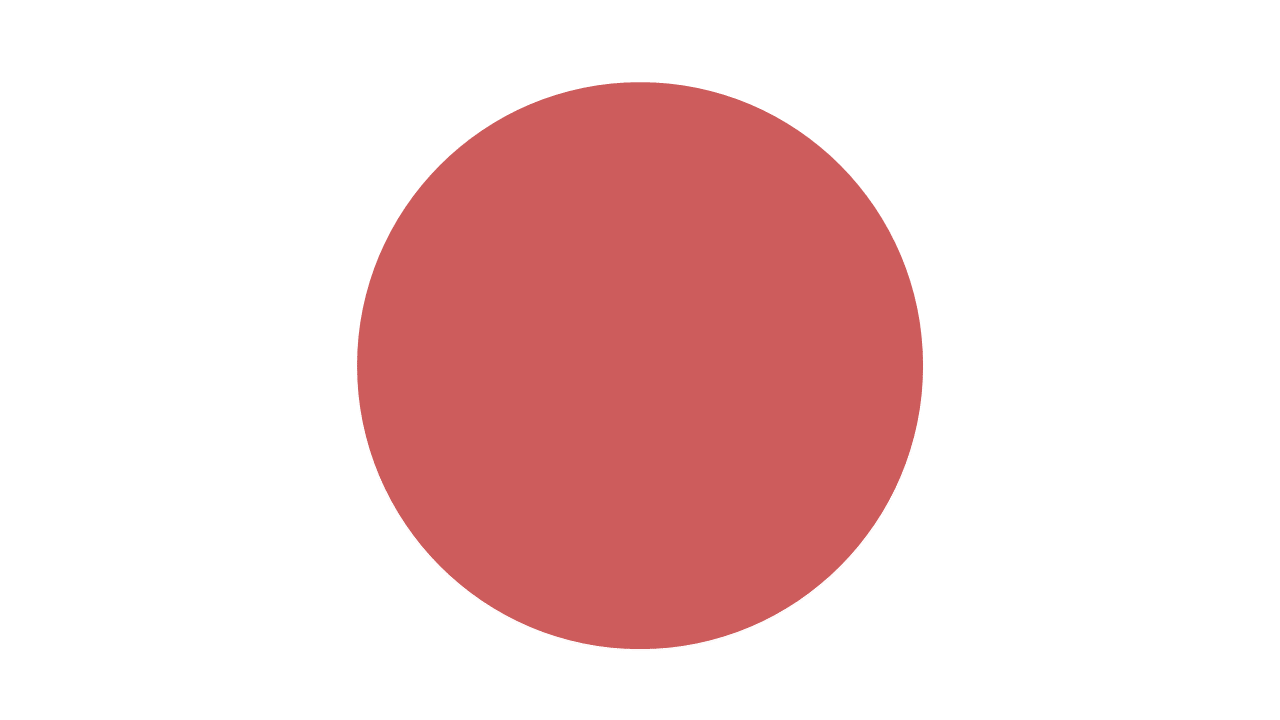}). Relevant indices are indicated, and lines' indices are underlined. Case studies on the 70-bus grid focus on buses indicated with \protect\includegraphics[height=0.5em]{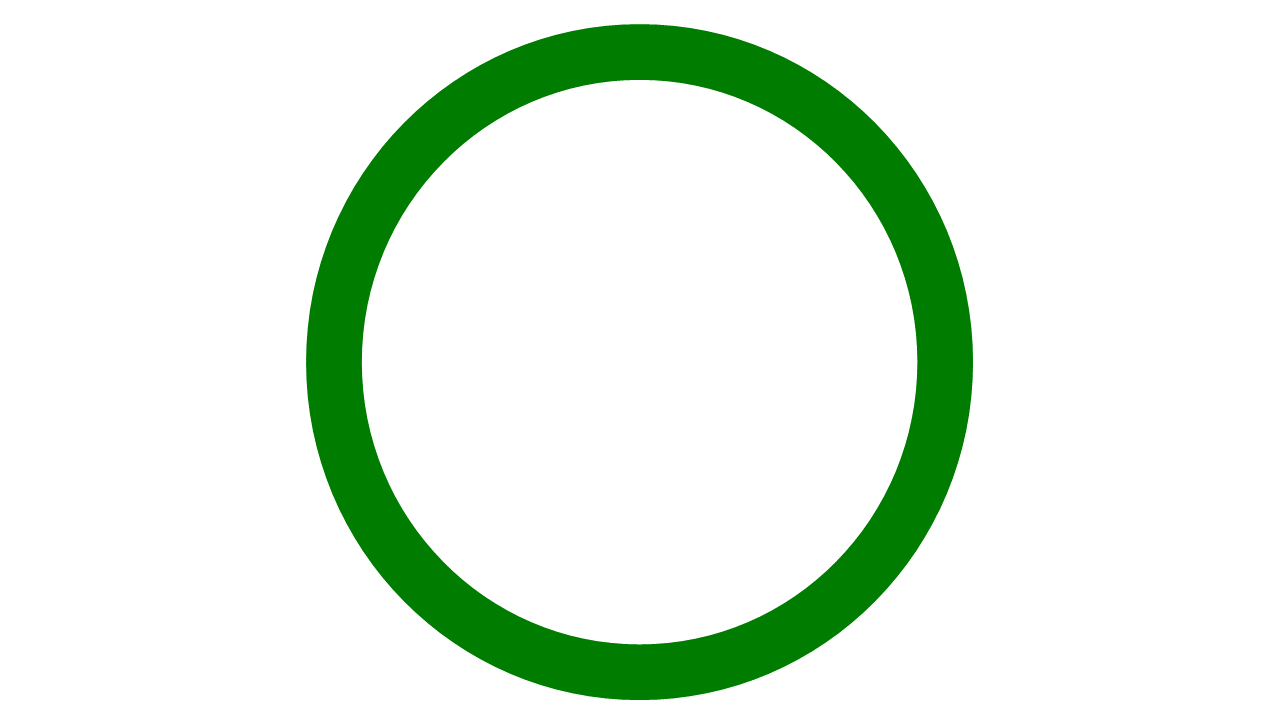}.\vspace{-1em}}
    \label{fig:testnetwork}
\end{figure}

We considered the 14-bus CIGRE MV distribution grid with PV and Wind distributed energy resources (DER) activated \cite{Barsali2014}, the 70-bus Oberrhein MV/LV sub-grid, and the whole 179-bus Oberrhein grid from \cite{Thurner2017}. The networks are presented in Figure \ref{fig:testnetwork}. The measurement locations for each network are shown in Fig. \ref{fig:testnetwork}. These measurements $\mathcal{M}$ either measure the power flow over lines or the voltages at buses and were assumed with different Gaussian noise, as further discussed.

For each network, $8640$ load samples were collected, equivalent to one year of hourly data. 
Each load scenario considers load levels of $24$ consecutive samples discretized hourly for all loads in the network. These load scenarios resulted from a Monte Carlo sampling on standard load profiles taken from \cite{Rudion2006}, considering a $15$\% uncertainty. For each sample, in each scenario, \textcolor{black}{assuming balanced systems,} the AC power flow computed the full \textit{true} state using PandaPower 2.9 \cite{Thurner2017} and Python 3.8. Subsequently, one sample's full \textit{true} state considered all loads and generators' active and reactive power levels, the bus voltage levels, phases, and line loadings. System operators do not have access to this full \textit{true} state; however, some key variables are provided by the measurements specified earlier. These observed variables were assumed corrupted with zero-mean Gaussian white noise at the measurement locations. Between $0.5$\% and $2$\% standard deviations were assumed for the voltage and current measurement noises, and between $1$\% and $5$\% for the active and reactive power measurement errors. Pseudomeasurements of power levels were considered at every (unobserved) bus using generic load and generation profiles taken from \cite{Rudion2006}.

The dataset was split into train, validation, and test sets, following an $80$/$10$/$10$ split. In supervised learning, the measurement vector $\bm{z}$ at the measurement locations mentioned above represents the input to the model, and the full state represents the label $\bm{y}$. 



Several baseline models were assumed as follows. The standard SE WLS algorithm \cite{Schweppe1970}, a standard ANN model trained with supervised learning, and the DSS\textsuperscript{2} model but trained with supervised learning (referenced with sup. DSS\textsuperscript{2}). The WLS algorithm from PandaPower 2.9 was used, and the deep learning models were implemented in Tensorflow 2.8 \cite{tf2.8}. The ANN was designed with $5$ layers of $32$ hidden values, $tanh$ activation functions and a Glorot normal initializer. \textcolor{black}{The code to reproduce the case studies of this paper can be accessed in GitHub \cite{HabibGit}.}



\vspace{-0.5em}
\subsection{Efficiency of the weakly-supervised learning}\label{sec:hyperparameter}
\begin{table}
\caption{Hyperparameter values of DSS\textsuperscript{2} trained for three power networks.}
\centering
\begin{tabular}{r|c c c}
\toprule
Parameter & {14-bus} & {70-bus} & {179-bus}        \\ 
 \midrule
Epochs                                                           & $630$                 & $540$                & $1020$                       \\ 
$\lambda$                                                                 & $0.8$                   & $0.8$                & $0.8$                        \\ 
$\alpha$                                                                  & $0.006$                 & $0.006 $             & $0.006$                      \\ 
batch size & $64$                   & $64$                & $64$                       \\ 
$d$                                                                       & $40$                    & $40$                 & $40$                         \\ 
{layers}                                                           & $3$                     & $3$                  & $3$                          \\ 
$T$                                                       & $7$         & $20$               & $25$                       \\ 
$r$                                                                       & $0.4$                   & $0.4$                & $0.4$                        \\ 
$\ell_2$                                                                  & $0.002$                 & $0.002$              & $0.002$ \\
\bottomrule
\end{tabular}
\label{param_tab}
\end{table}

This section investigates the efficiency of the weakly supervised learning DSS\textsuperscript{2} approach and hyperparameters that can impact the state estimation accuracy.
The hyperparameters penalization factor 
$\lambda=\lambda_0=\lambda_1=\lambda_2$, batch size, dropout rate $r$, $\ell_2$-regularizer, and the number of iteration $T$ were fixed. 
A grid search tuned the hyperparameters learning rate within the ranges $\alpha \in \{0.001, 0.002,\cdots,0.009,0.01\}$, layer dimension $d\in\{8,16,24,32,40,48\}$, and layers number  $\in \{2,3,4,5\}$. The selected hyperparameter values are in Table \ref{param_tab} for each network.

\begin{figure}
\begin{subfigure}[t]{0.18\textwidth}
\begin{tikzpicture}
\begin{axis}[
        axis on top,
        width=\textwidth,
        scale only axis,
        enlargelimits=false, 
        ytick={0.005,0.01,0.015,0.02,0.025},     
        xtick={0, 200, 400, 600, 800}, 
        ylabel={RMSE [-]},
        xlabel={Training epochs},
        xmin=-20,
        xmax=800,
        ymin=0.005,
        ymax=0.026,
        y label style={at={(axis description cs:-0.15,.5)},anchor=south}, 
        ]        
	\addplot graphics[xmin=0,ymin=0.005,xmax=800,ymax=0.026] {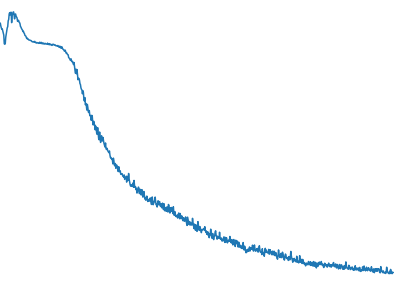};
  \end{axis}
\end{tikzpicture}
   \caption{}
   \label{v_val}
\end{subfigure} \hspace{0.05\textwidth}
\begin{subfigure}[t]{0.18\textwidth} 
\begin{tikzpicture}
\begin{axis}[
        axis on top,
        width=\textwidth,
        scale only axis,
        enlargelimits=false, 
        ytick={0,10,20,30}, 
        xtick={0, 200, 400, 600, 800},         
        ylabel={RMSE [\%]},
        xlabel={Training epochs},
        xmin=-20,
        xmax=800,
        ymin=0,
        ymax=32,
        y label style={at={(axis description cs:-0.15,.5)},anchor=south}, 
        ]        
	\addplot graphics[xmin=0,ymin=0,xmax=800,ymax=30] {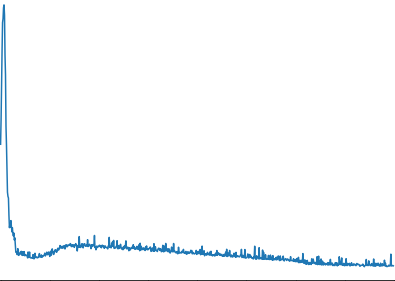};
  \end{axis}
\end{tikzpicture}
   \caption{}
   \label{load_val}
\end{subfigure}
\caption{Validation RMSEs of voltages (a) and line loading (b) during training. \vspace{-1em}}
\label{val}
\end{figure}

The efficiency of the learning approach is shown in Figures \ref{v_val} and \ref{load_val} when training on the $14$-bus network. The voltage and line loading estimation RMSE slowly decreased at each epoch, showing a learning curve through the power flow equations and using only the noisy measurements and pseudomeasurements. When training in weakly supervision, DSS\textsuperscript{2} learned to minimize the different objectives using \textit{noisy} measurements as 'reference values'. However, the computational time to train DSS\textsuperscript{2} in weakly supervision was lower than to train in supervision. 

\vspace{-0.5em}
\subsection{Trade-off between accuracy and available labels} \label{sec:accuracy}

\begin{table*}
\caption{Mean (standard deviation in parentheses) values of performance metrics in \textit{default} conditions. WLS* is the WLS algorithm without flow measurements, and WLS** with increased tolerance. The bold font shows the best model or algorithm for each metric.}
\centering
\smaller
\begin{tabular}{r|cccc|ccc|cc}
\toprule
& \multicolumn{4}{c|}{14-bus CIGRE} & \multicolumn{3}{c|}{70-bus Oberrhein} & \multicolumn{2}{c}{179-bus Oberrhein} \\
\midrule
Performance metric \textbackslash \, Model & {WLS}   & {ANN} & {sup. DSS\textsuperscript{2}} & {DSS\textsuperscript{2}} &  {WLS} & {WLS*} & {DSS\textsuperscript{2}} & {WLS**} & {DSS\textsuperscript{2}}\\ 
\midrule
{Voltage RMSE}  [$10^{-3}$]                  & $9.9 \, (0.45) $          & $2.7 \,(0.17)$             & $\mathbf{2.5\, (0.11)}$     & $3.4\, (0.18)$   & $31\, (0.96)$   & $5.9 \,(0.18)$            & $\mathbf{1.5\, (0.01)}$    & $9.9\, (0.3)$      & $\mathbf{2.3\, (0.01)}$      \\ 
{Line loading RMSE} [\%]             & $\mathbf{3.4\, (0.05)}$   & $42 \,(0.83)$         & $12.7\, (0.17)$       & $3.8 \,(0.05)$    & $17 \,(0.9)$         & $15 \,(0.8)$     & $\mathbf{2.3\, (0.01)}$     & $5.9\, (0.4)$       & $\mathbf{3.4\, (0.01)} $        \\
{Line \& trafos loading RMSE} [\%]                      & $\mathbf{4.6\, (0.06)}$  & $39 \,(0.5)$         & $14\, (0.15)$        & $8 \,(0.06)$     & $39\, (2.7)$         & $24 \,(1.67)$             & $\bm{2.5\, (0.01)}$  & $4.2\, (0.2)$        & $\mathbf{3.5\, (0.01)}$       \\ 
{Convergence} [\%]               & $100$              & $100$            & $100$           & $100$      & $25$           & $100 $     & $100$   & $53$        & $\mathbf{100}$             \\ 
{Computational time} [ms]                               & $86\, (41)$           & $\mathbf{3.5\, (0.73)}$ & $4.7\, (1.4)$        & $5.5\, (4.5)$    & $123\, (31)$      & $161\, (40)$        & $\mathbf{26 \,(10.4)}$        & $1212\, (424)$   & $\mathbf{58 \,(26.7)}$          \\ 
\bottomrule
\end{tabular}
\label{tab:results}
\end{table*}

\begin{figure}
\begin{subfigure}{0.35\textwidth}
\begin{tikzpicture}
\begin{axis}[
        axis on top,
        width=\textwidth,
        height = 7em,
        scale only axis,
        enlargelimits=false, 
        ytick={0, 0.005,0.01,0.015,0.02,0.02},     
        xtick={0,1,2,3,4,5,6,7,8,9,10,11,12,13,14 }, 
        ylabel={RMSE [-]},
        xlabel={Bus index},
        xmin=-1,
        xmax=15,
        ymin=0,
        ymax=0.021,
        y label style={at={(axis description cs:-0.1,.5)},anchor=south}, 
        ]        
	\addplot graphics[xmin=-1.1,ymin=0,xmax=15.1,ymax=0.021] {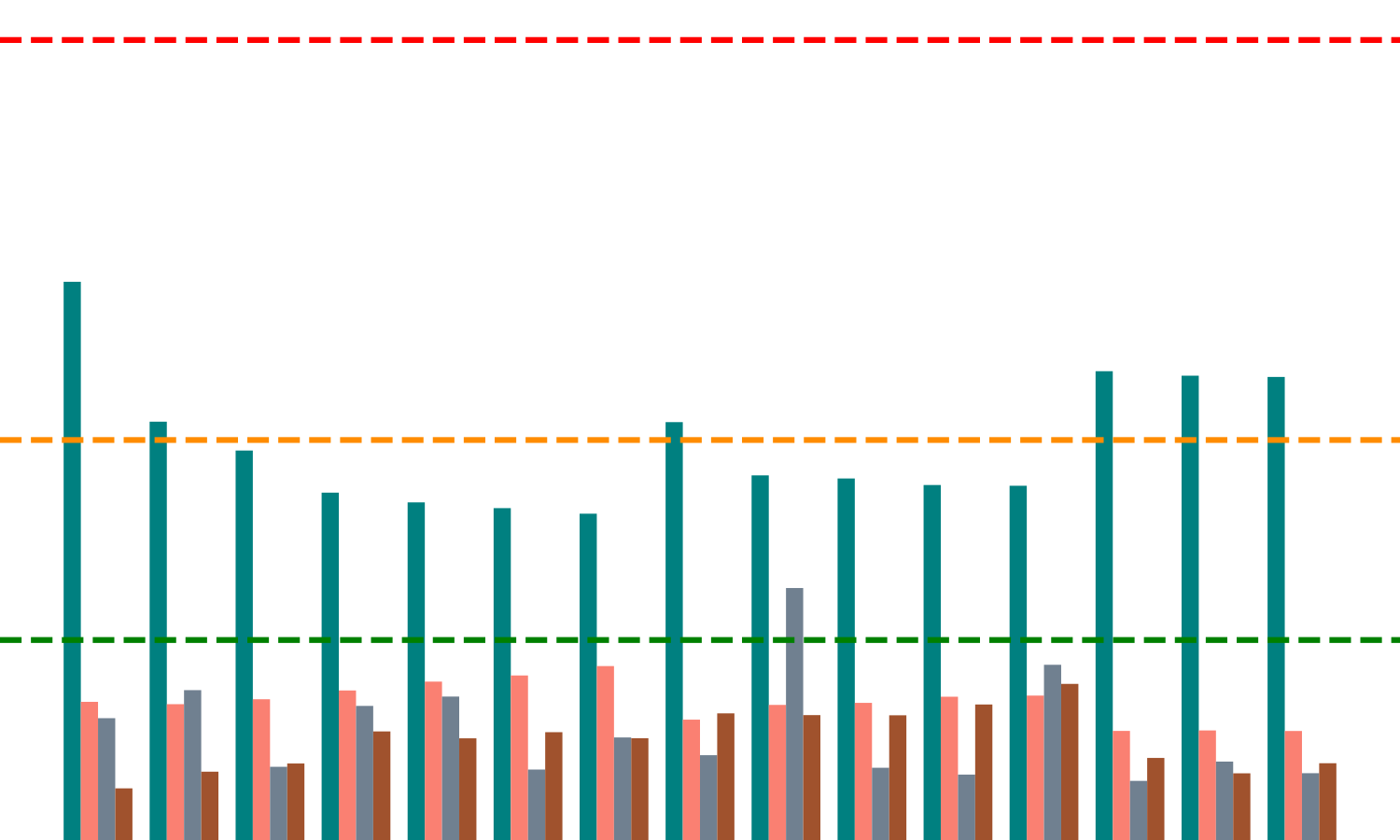};
  \end{axis}
\end{tikzpicture}
   \caption{}
\label{cigre_cs1_vbus}
\end{subfigure} 
\begin{subfigure}{0.35\textwidth} 
\begin{tikzpicture}
\begin{axis}[
        axis on top,
        width=\textwidth,
        height = 7em,
        scale only axis,
        enlargelimits=false, 
        ytick={0,10,20,30,40}, 
        xtick={0,1,2,3,4,5,6,7,8,9,10,11,12,13},     
        ylabel={RMSE [\%]},
        xlabel={Line index},
        xmin=-1,
        xmax=14,
        ymin=0,
        ymax=40,
        y label style={at={(axis description cs: -0.1,.5)},anchor=south}, 
        ]        
	\addplot graphics[xmin=-1.1,ymin=0,xmax=14.1,ymax=40] {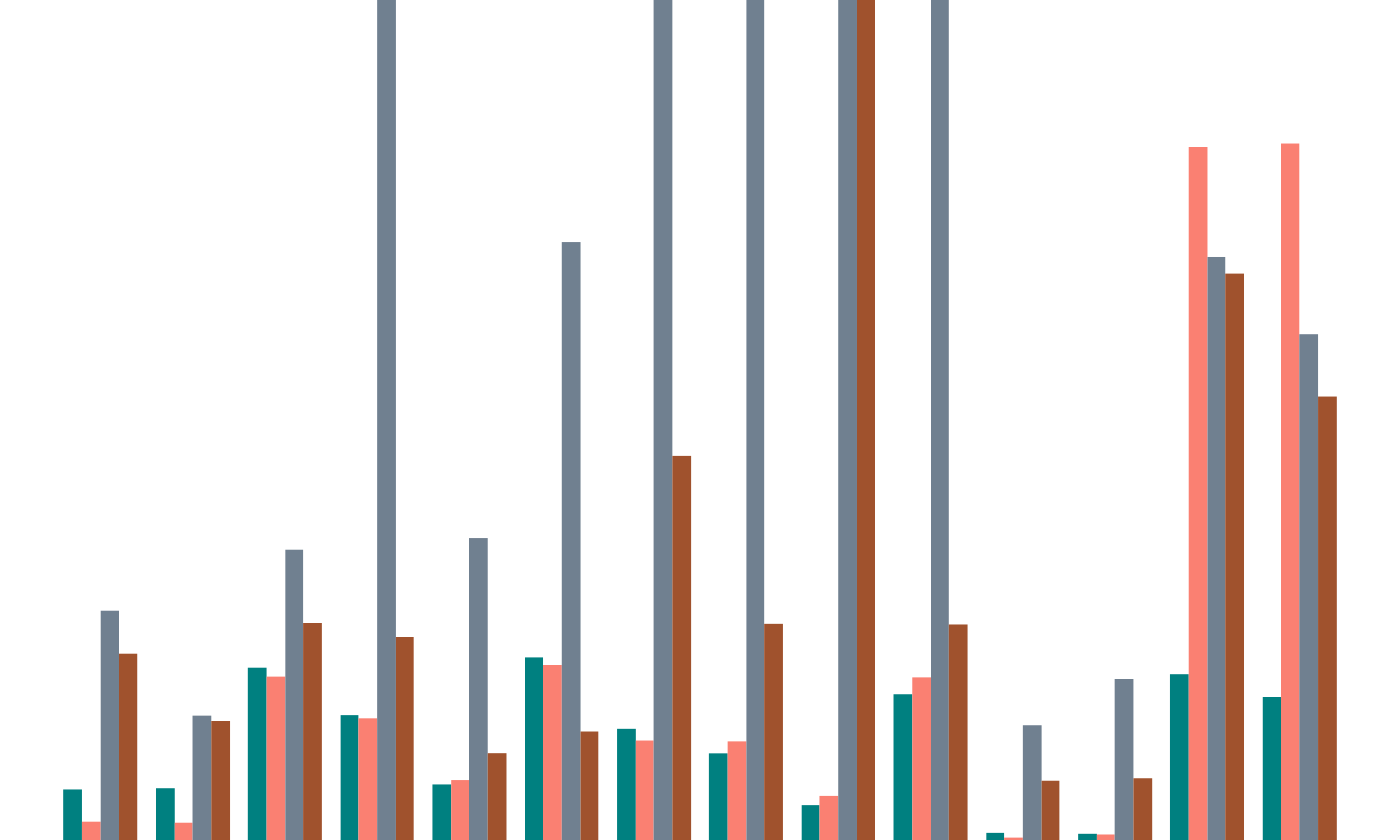};
  \end{axis}
\end{tikzpicture}
   \caption{}
\label{cigre_cs1_load}
\end{subfigure}
\caption{Comparing RMSE of estimating (a) voltage levels and (b) line loadings in the 14-bus network between WLS (\protect\includegraphics[height=0.5em]{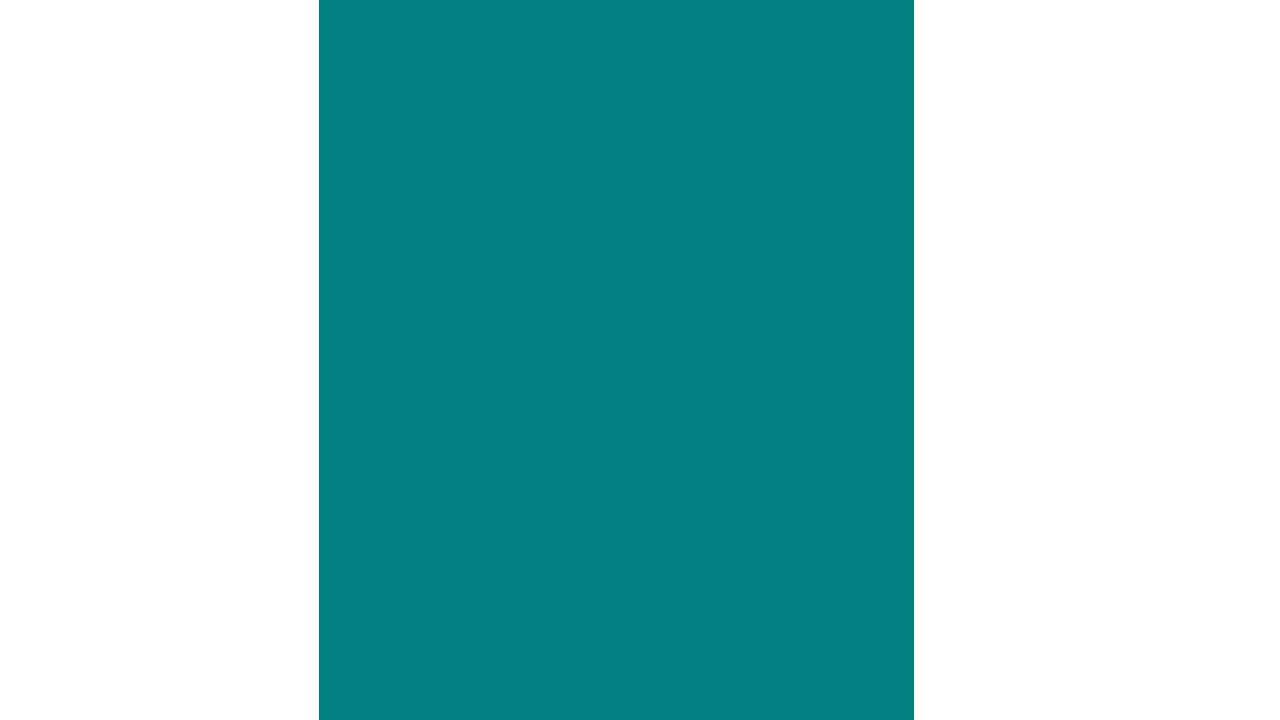}), DSS\textsuperscript{2} (\protect\includegraphics[height=0.5em]{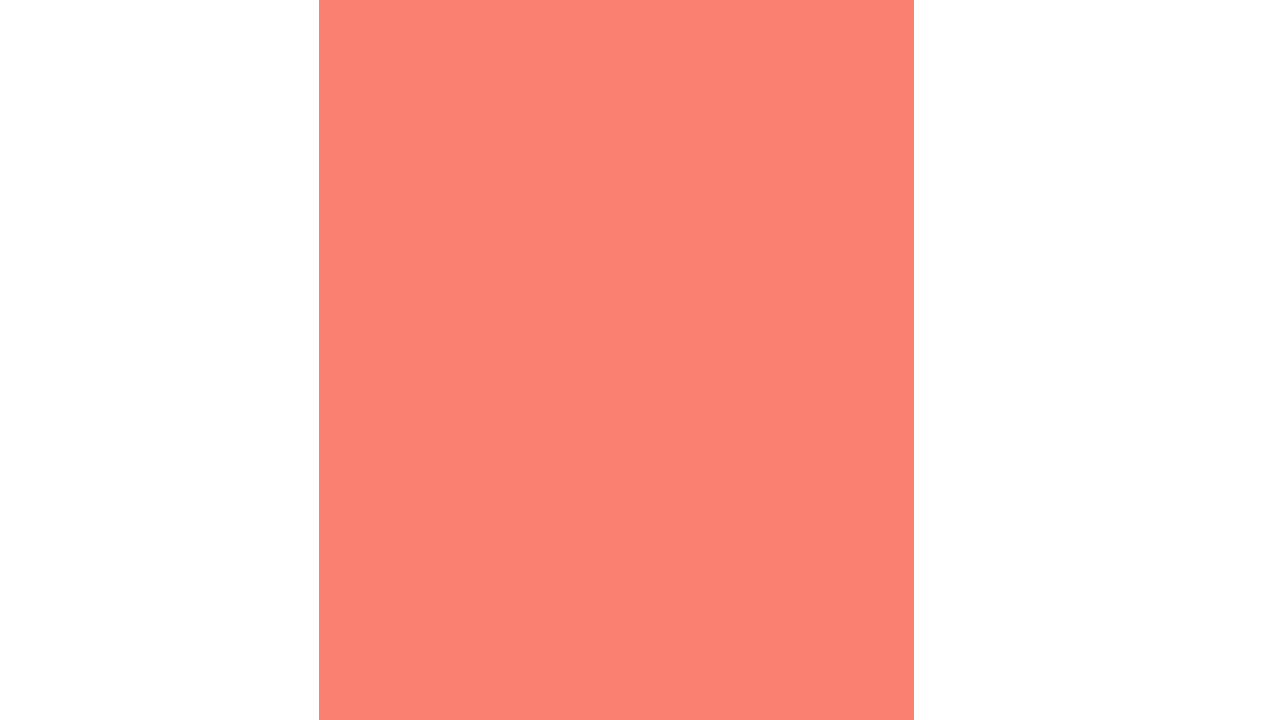}), FFNN (\protect\includegraphics[height=0.5em]{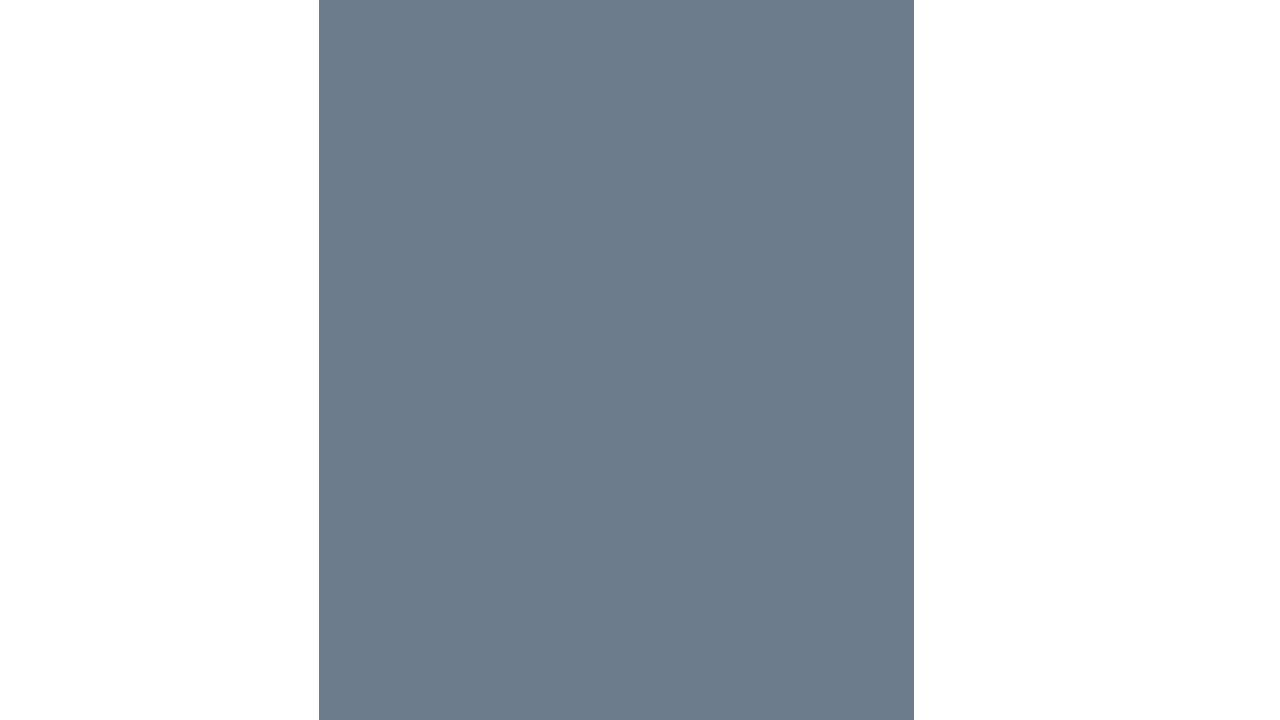}) and sup. DSS\textsuperscript{2} (\protect\includegraphics[height=0.5em]{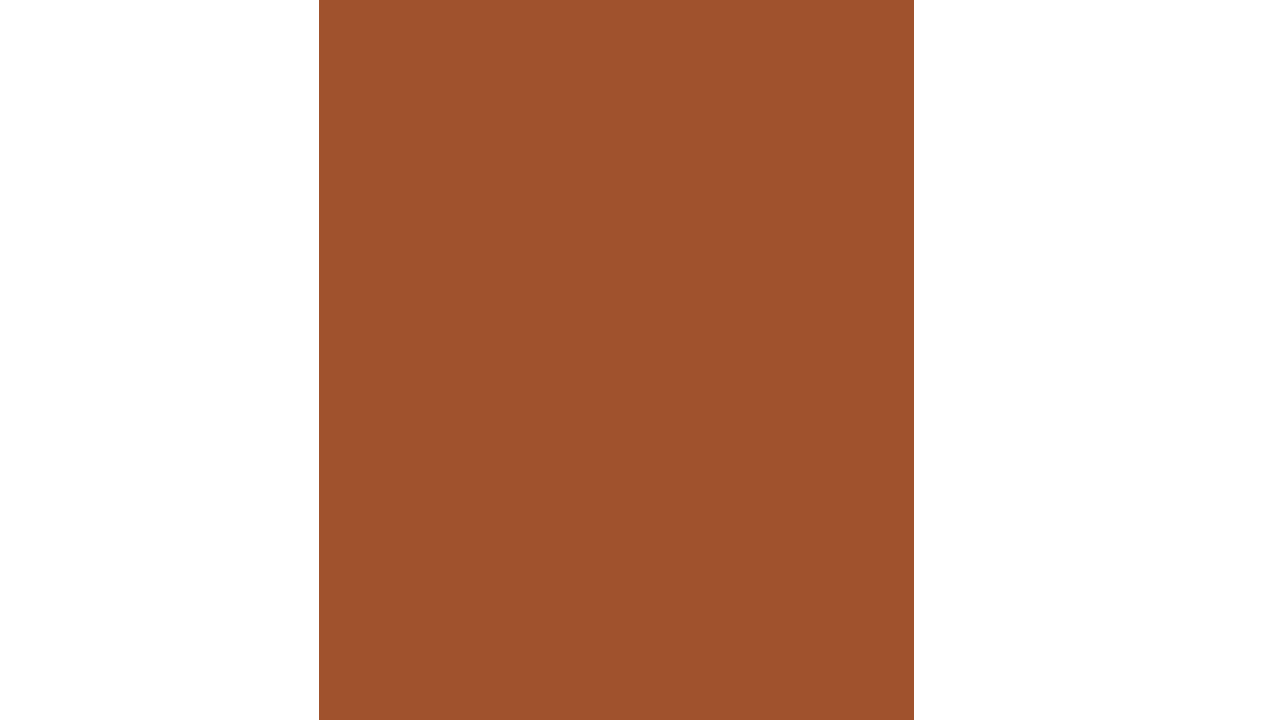}). In (a), the dashed lines show acceptable state estimators based on the standard deviation. Below the dashed green is acceptable, and above red is unacceptable. In (b), indexes $12$ and $13$ are transformers. \vspace{-1em}}
\label{cigre_cs1}
\end{figure}

This case study investigates the performance of the proposed weakly supervised DSS\textsuperscript{2} model on the $14$-bus system compared to three baselines. The second column in Table \ref{tab:results} summarizes the results.\footnote{\textcolor{black}{Observability in the large networks is reached artificially by using pseudomeasurements in unobservable parts of the networks.}}  


The RMSE for voltages of the proposed weakly supervised DSS\textsuperscript{2} was three times lower than the WLS, $2.5$\% versus $9.9$\%. In more detail, Figure \ref{cigre_cs1_vbus} shows the voltage estimation RMSE per bus. The RMSE was lower than the $0.5\%$ threshold for all buses, showing successful learning from voltage measurement data while handling measurements' noise. The difference in RMSE between the observed (buses $1$,$8$, and $12$) and the unobserved buses are small, showing the capability of our DSS\textsuperscript{2} model to extrapolate to all buses. The supervised models (ANN, sup. DSS\textsuperscript{2}) estimated the voltage more accurately, as expected, as they learned from the ideal \textit{true} voltage data having an unfair, impractical advantage. 

The RMSE of line loading of the weakly supervised DSS\textsuperscript{2} reaches performances equivalent to the WLS, outperforming the supervised models by a wide margin according to Table \ref{tab:results}. This observation offered insights. Supervised models poorly estimated \textit{indirect} values such as the line loading that were calculated using the power flow equations. The models only outputted the state variables and supervised models poorly considered the \textit{coupling} of the state variables in the estimations of line loading. However, the weakly supervised model learned directly through the power flow equations about the coupling with the effect of estimating line loading more accurately. In more detail, Figure \ref{cigre_cs1_load} shows the loading estimation error per line. 
The weakly supervised DSS\textsuperscript{2} model had a very high accuracy on measured lines (lines $0$ and $10$) and their extension (lines $1$ and $11$). However, there was a clear drop in performance for the estimation of transformers' loading, shown at indexes $12$ and $13$. The simple modelling of transformers or slack may have led to this reduced accuracy as the transformers and lines were considered in the same class of models. 
As a result of this simple modelling, the H2MGNN considered the same mapping for these components, which may have reduced the accuracy of transformer estimations.

\vspace{-0.5em}
\subsection{Convergence, accuracy and computation speed in large networks}\label{sec:scalability}

This case study investigates the performance of the proposed DSS\textsuperscript{2} compared to the WLS in larger networks, the $70$-bus and $179$-bus networks, along three performance criteria: the convergence rate, the accuracy, and the computational time. The 2\textsuperscript{nd} and 3\textsuperscript{rd} columns in Table \ref{tab:results} summarize the results.

When analysing the convergence rate, the DSS\textsuperscript{2} always converges, and the WLS never converges in the $179$-bus network. 
The WLS  was unstable in this large and noisy network, leading to these poor convergence rates. WLS' convergence issues with noisy measurements in large systems is already well-known \cite{Ahmad2018,fot22}. Many noisy measurements constrain the Newton-Raphson solver and can lead to divergence. More specifically, the WLS had issues in handling flow measurements. In response to these issues and to compare the accuracy and computational times of DSS\textsuperscript{2} with WLS, only voltage measurements and pseudomeasurements were used in WLS to increase the convergence rate (WLS* in the table). This increased the convergence rate in the $70$-bus system but did not increase the convergence of the WLS in the $179$-bus. Therefore, in the $179$-bus system, the tolerance of the Newton-Raphson iterative process and the number of iterations were increased (WLS** in the table). Increasing these parameters increased the convergence rate at the cost of lower accuracy and slower processing.

\begin{figure}
\begin{subfigure}[b]{0.35\textwidth}
\begin{tikzpicture}
\begin{axis}[
        axis on top,
        width=\textwidth,
        height = 7em,
        scale only axis,
        enlargelimits=false, 
        ytick={0.98, 1.00, 1.02, 1.04, 1.06},    
        xtick={0,30,60,90,120}, 
        ylabel={Voltage [pu]},
        xlabel={Time [h]},
        xmin=0,
        xmax=120,
        ymin=0.98,
        ymax=1.06,
        y label style={at={(axis description cs:-0.15,.5)},anchor=south}, 
        ]        
	\addplot graphics[xmin=-5,ymin=0.95,xmax=125,ymax=1.05] {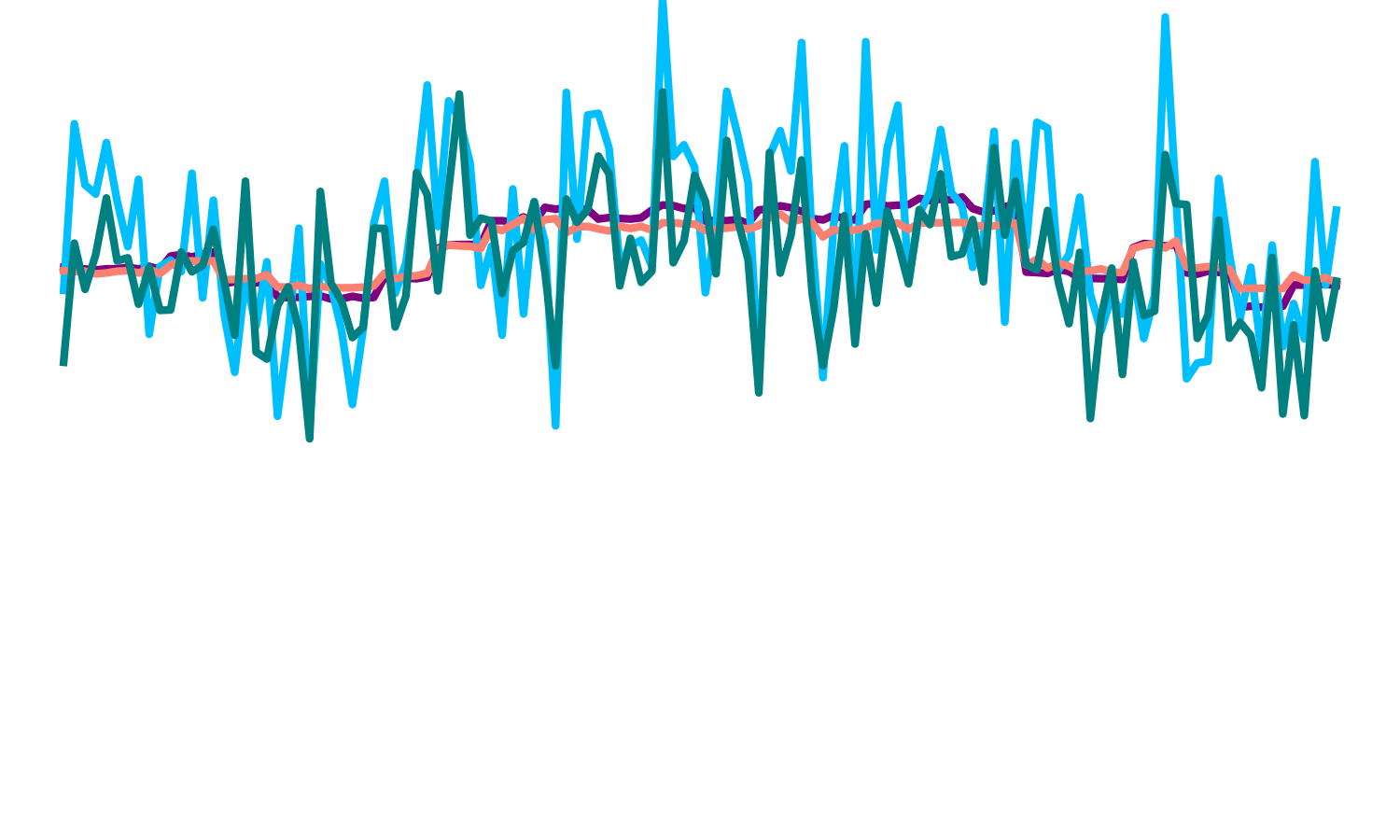};
  \end{axis}
\end{tikzpicture}
   \caption{\centering { } }
\label{ober_cs1_bus34}
\end{subfigure} 
\begin{subfigure}[b]{0.35\textwidth} 
\begin{tikzpicture}
\begin{axis}[
        axis on top,
        width=\textwidth,
        height = 7em,
        scale only axis,
        enlargelimits=false, 
        ytick={0.99, 1, 1.01, 1.02, 1.03},     
        xtick={0,30,60,90,120}, 
        ylabel={Voltage [pu]},
        xlabel={Time [h]},
        xmin=0,
        xmax=120,
        ymin=0.99,
        ymax=1.03,
        y label style={at={(axis description cs:-0.15,.5)},anchor=south}, 
        ]        
	\addplot graphics[xmin=-5,ymin=0.95,xmax=125,ymax=1.05]{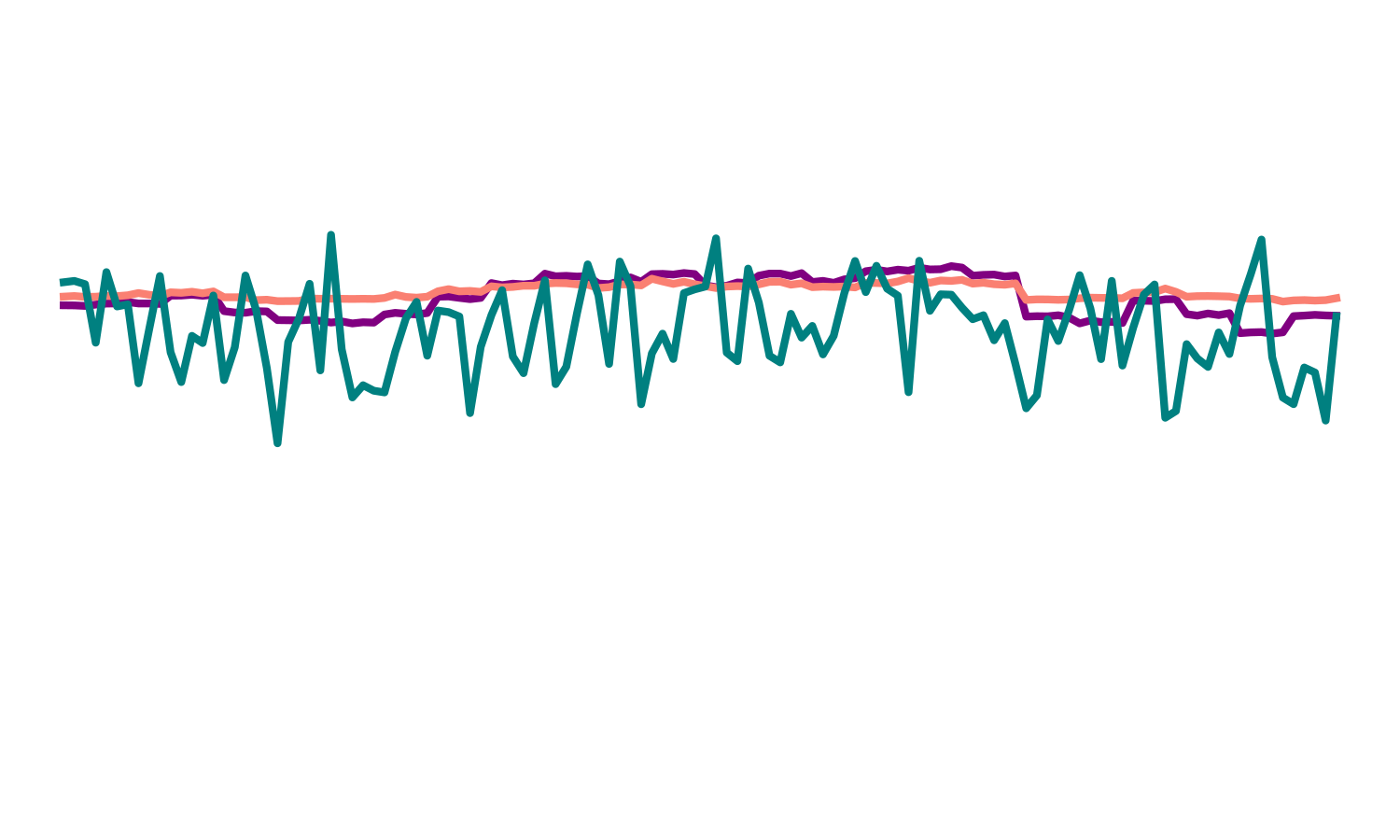};
  \end{axis}
\end{tikzpicture}
   \caption{}
\label{ober_cs1_bus223}
\end{subfigure}
\caption{Estimation of the voltage level at (a) bus $34$ and (b) bus $223$ of the $70$-bus network under normal conditions and across the sampling period, using WLS (\protect\includegraphics[height=0.5em]{green.png}), and DSS\textsuperscript{2} (\protect\includegraphics[height=0.5em]{orange.png}). True voltage (\protect\includegraphics[height=0.5em]{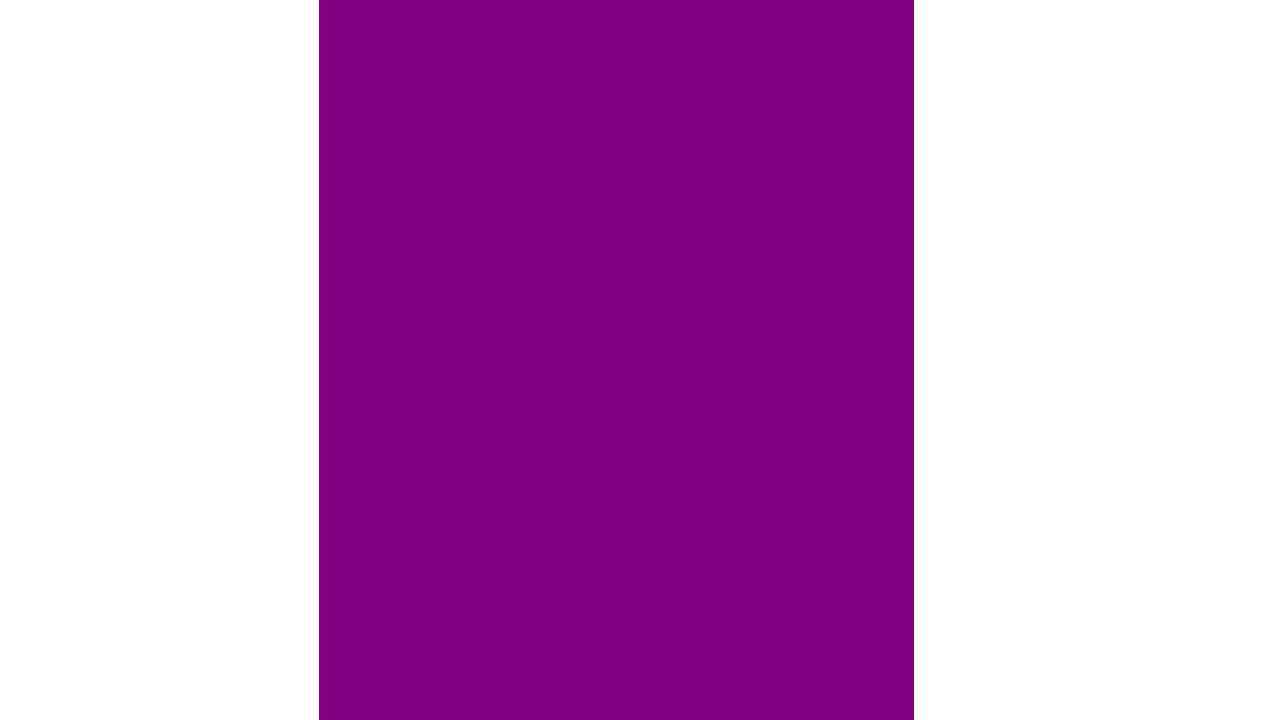}) and measurement (\protect\includegraphics[height=0.5em]{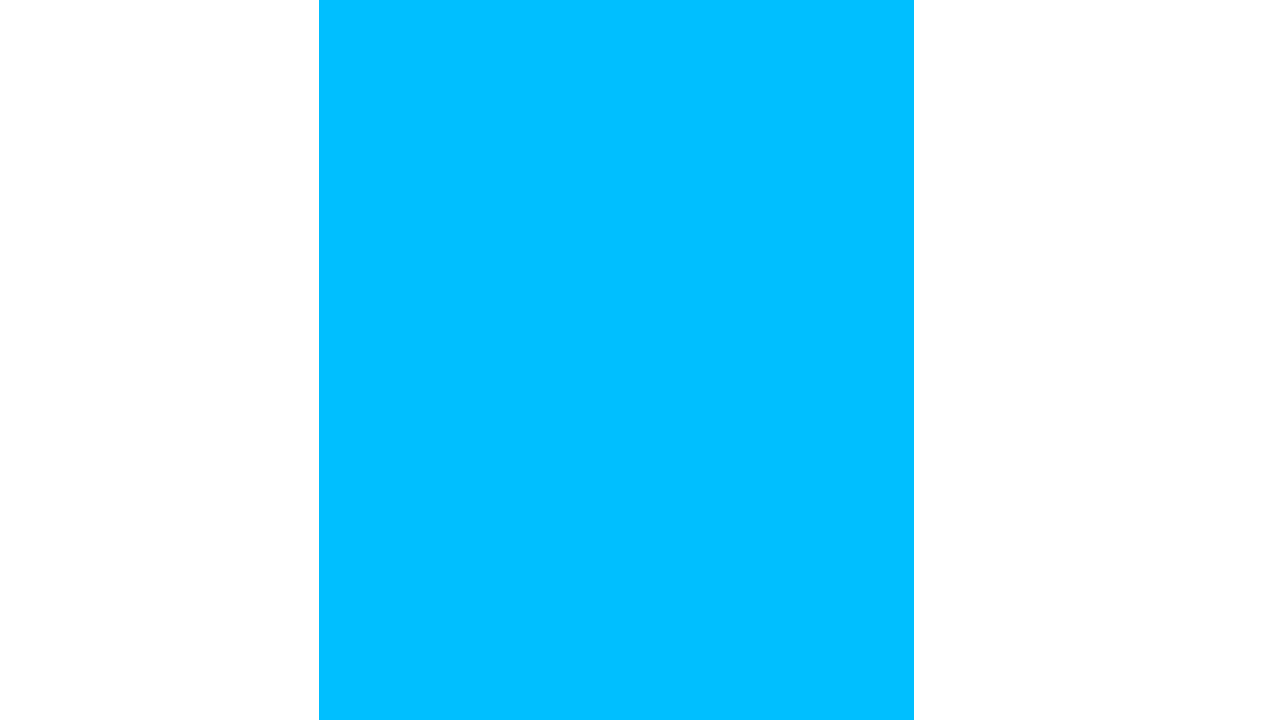}) as references. \vspace{-1em}}
\label{ober_cs1_bus}
\end{figure}

When analysing the accuracy, a key advantage of the DSS\textsuperscript{2} becomes visible. DSS\textsuperscript{2} outperformed the WLS in every metric in the two larger networks. The models based on GNN, such as DSS\textsuperscript{2}, learn from local operations (in the neighbourhood of buses) and extrapolate to other locations (to other neighbourhoods of buses). Therefore, the more buses and lines in the network, the more local operations to learn from that can further enhance the model's accuracy. Also, these networks have more static loads and less DER than the $14$-bus network, so the variation of voltage and line loading was smaller, and the estimated values from the DSS\textsuperscript{2} become more accurate. Figures \ref{ober_cs1_bus34} and \ref{ober_cs1_bus223} compare the estimated voltage levels through a sampling period in the $70$-bus system for the measured bus $34$ and the unmeasured remote bus $223$, respectively. The accuracy of the DSS\textsuperscript{2} model estimating the voltage in measured nodes through noisy measurements was high. However, the model lacked generalizability when estimating voltage in remote, unmeasured nodes. 

When analysing the computation times, in the last row of Table \ref{tab:results}, the DSS\textsuperscript{2} increasingly outperformed WLS for larger networks. The computational time of the WLS and the DSS\textsuperscript{2} increased from the $70$-bus network to the $179$-bus network by factors of $10$ and $2$, respectively. The DSS\textsuperscript{2} scaled to a larger network $5$-fold better than the WLS algorithm. The WLS needed more iterations for this larger system until the Newton-Raphson converged, although the tolerance was increased, which typically decreased the computational times. The DSS\textsuperscript{2} also showed a lower variance in the computational times as it is not based on an iterative algorithm.

\vspace{-0.5em}
\subsection{Measurement noise} \label{sec:noise}

\begin{figure}
\begin{subfigure}[t]{0.18\textwidth}
\begin{tikzpicture}
\begin{axis}[
        axis on top,
        width=\textwidth,
        height = 7em,
        scale only axis,
        enlargelimits=false, 
        xticklabel style={font=\footnotesize},        
        ytick={0, 0.005,0.01,0.015,0.02},     
        xtick={1.5, 4, 6.5}, 
        xticklabels = {\textit{default}, \textit{high}, \textit{low}}, 
        ylabel={RMSE [-]},
        xlabel={Noise},
        xmin=0,
        xmax=8,
        ymin=0,
        ymax=0.02,
        y label style={at={(axis description cs:-0.15,.5)},anchor=south}, 
        ]        
	\addplot graphics[xmin=0,ymin=0,xmax=8,ymax=0.02] {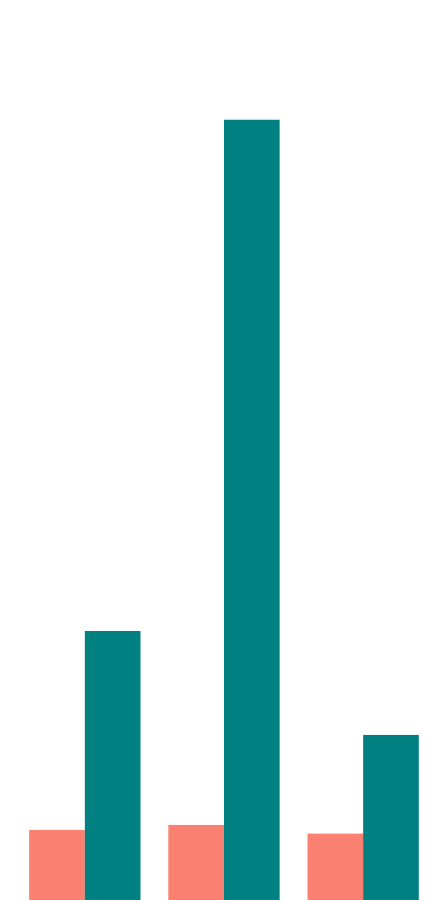};
  \end{axis}
\end{tikzpicture}
   \caption{}
\label{ober_v_noise}
\end{subfigure}\hspace{0.05\textwidth}
\begin{subfigure}[t]{0.18\textwidth} 
\begin{tikzpicture}
\begin{axis}[
        axis on top,
        width=\textwidth,
        height = 7em,
        scale only axis,
        enlargelimits=false, 
        ytick={0, 25, 50, 75, 100}, 
        xticklabel style={font=\footnotesize}, 
        xtick={1.5, 4, 6.5}, 
        xticklabels = {\textit{default}, \textit{high}, \textit{low}}, 
        ylabel={RMSE [\%]},
        xlabel={Noise},
        xmin=0,
        xmax=8,
        ymin=0,
        ymax=110,
        y label style={at={(axis description cs: -0.15,.5)},anchor=south}, 
        ]        
	\addplot graphics[xmin=0,ymin=0,xmax=8,ymax=110] {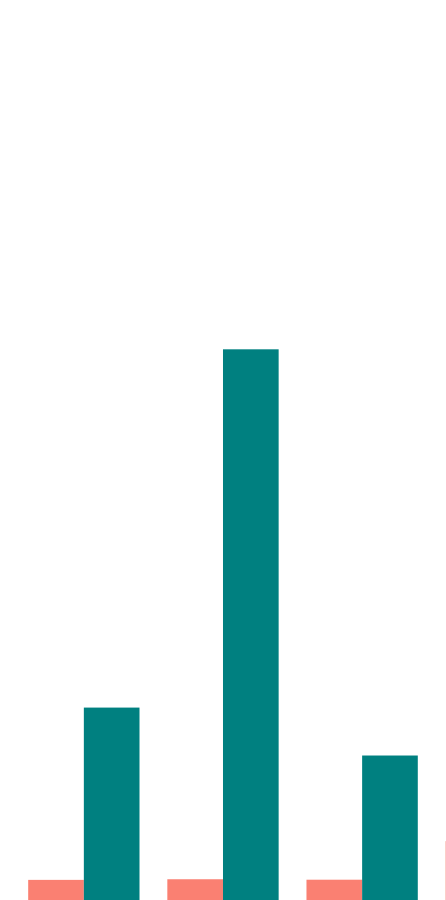};
  \end{axis}
\end{tikzpicture}
   \caption{}
\label{ober_load_noise}
\end{subfigure}
\caption{Comparison of RMSE for the estimation of (a) voltage level and (b) line loading between the WLS (\protect\includegraphics[height=0.5em]{green.png}) and DSS\textsuperscript{2} (\protect\includegraphics[height=0.5em]{orange.png}) in the 70-bus network; considering different levels of measurement noise.\vspace{-2em}}

\label{ober_rmse_noise}
\end{figure}

This case study compares the robustness to measurement noise of the DSS\textsuperscript{2} to the WLS in the 70-bus network. The level of measurement noise refers to the standard deviation $\sigma_i$ of the Gaussian noise added to the measurements. Three different levels of noise were considered. The \textit{default} level had $1$\% noise on ideal measurements of voltage and current, and $2$\% noise on the ideal measurements of active and reactive power; the \textit{low} level had $0.5$\% and $1$\% noise, and the \textit{high} level $3$\% and $5$\%, respectively. 

\begin{figure}
\begin{tikzpicture}
\begin{axis}[
        axis on top,
        width=0.35\textwidth,
        height = 7em,
        scale only axis,
        enlargelimits=false, 
        ytick={0.95, 1, 1.05 },     
        xtick={0,30,60,90,120}, 
        ylabel={Voltage [pu]},
        xlabel={Time [h]},
        xmin=0,
        xmax=120,
        ymin=0.94,
        ymax=1.06,
        y label style={at={(axis description cs:-0.15,.5)},anchor=south}, 
        ]        
	\addplot graphics[xmin=-5,ymin=0.95,xmax=125,ymax=1.05]{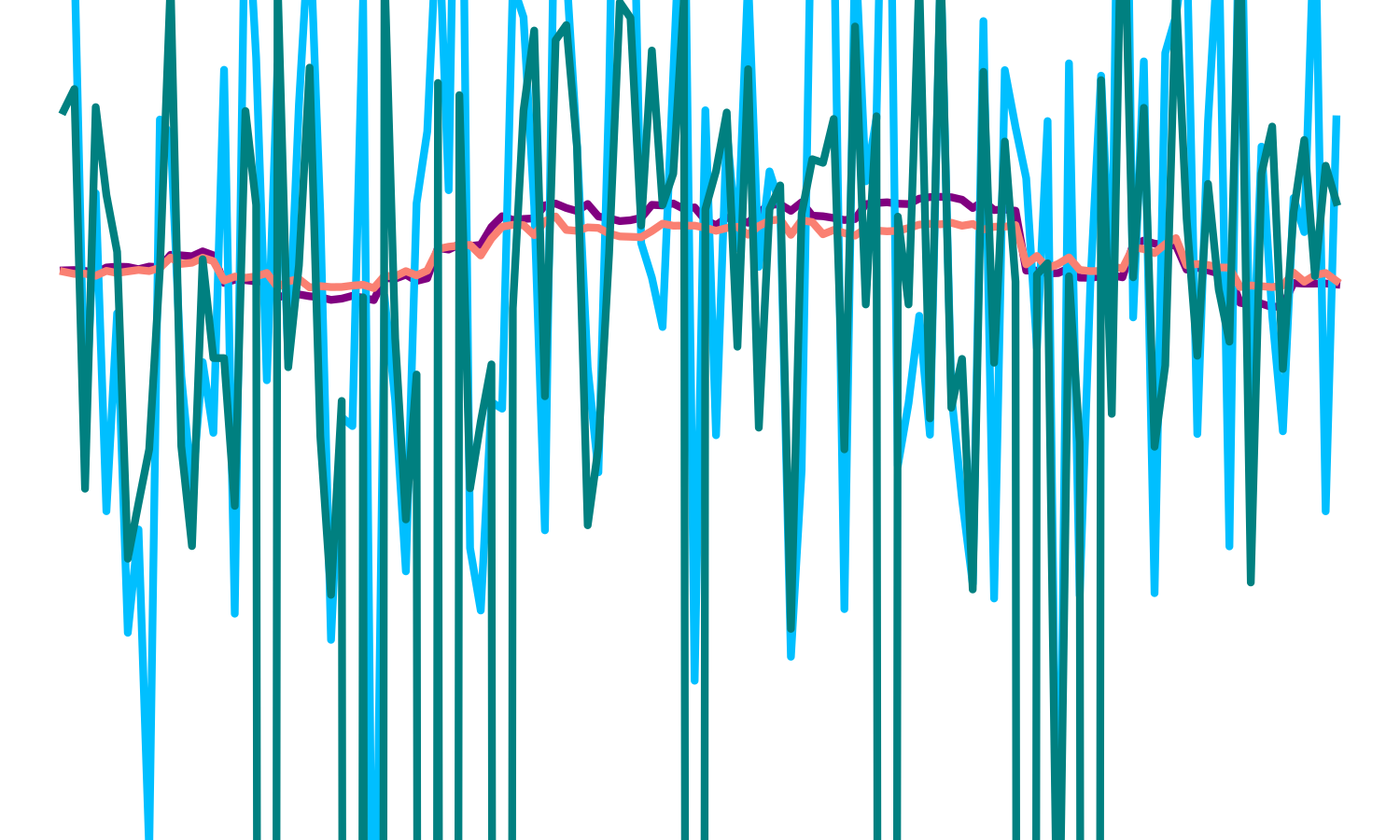};
  \end{axis}
\end{tikzpicture}
\caption{Estimation of the voltage level at (a) bus $34$ of the 70-bus network under \textit{high} noise level and across the sampling period, using WLS (\protect\includegraphics[height=0.5em]{green.png}), and DSS\textsuperscript{2} (\protect\includegraphics[height=0.5em]{orange.png}). True voltage (\protect\includegraphics[height=0.5em]{purple.png}) and measurement (\protect\includegraphics[height=0.5em]{lightblue.png}) as reference. \vspace{-1em} }
\label{ober_cs2_bus34}
\end{figure}

At \textit{high} noise, Fig. \ref{ober_rmse_noise} shows the RMSE of the DSS\textsuperscript{2} was more than $10$ times better than that of the WLS showing significantly higher robustness of DSS\textsuperscript{2}. DSS\textsuperscript{2} had a similarly high accuracy at \textit{low} and \textit{high} noise as in the \textit{default} noise level. DSS\textsuperscript{2} learned to process many noisy signals with different standard deviations within the high noise level ranges and GNN structures. The dropout step during training improved the capability of the DSS\textsuperscript{2} model to handle stochasticity, including noise. Fig. \ref{ober_cs2_bus34} compares the voltage level estimation at \textit{high} measurement noise for the bus $34$. The DSS\textsuperscript{2} successfully cancelled the increased noise, whereas the WLS algorithm struggled to stay accurate. 

\subsection{Missing and erroneous measurements}\label{sec:error}
This case study investigates the impact of missing and erroneous measurements on the DSS\textsuperscript{2} and the WLS algorithm at the $70$-bus network. Case (i) assumed a missing voltage measurement on bus $39$ that was naively replaced with their historical mean value. Case (ii) assumed erroneous voltage measurements on buses $39$, $58$ and $80$, and erroneous active power flow measurements in lines $162$ and $165$ with a higher deviation from the \textit{true} state values than the expected (standard) deviation. Case (iii) assumed missing voltage measurements on  buses $34$, $39$ and $80$ and erroneous voltage measurements on bus $58$. 

\begin{figure}
\begin{subfigure}[t]{0.18\textwidth}
\begin{tikzpicture}
\begin{axis}[
        axis on top,
        width=\textwidth,
        height = 7em,
        scale only axis,
        enlargelimits=false,   
        xticklabel style={rotate = 0, align=center, font=\footnotesize, text width = 1.1cm, shift={(0,0 |- {axis description cs:0,0})}},
        ytick={0, 0.005,0.01,0.015,0.02},     
        xtick={1.5, 4, 6.5}, 
        xticklabels =  {(i) missing, (ii) \\ erron., (iii) erron. \& missing}, 
        ylabel={RMSE [-]},
        xmin=0,
        xmax=8,
        ymin=0,
        ymax=0.02,
        y label style={at={(axis description cs:-0.15,.5)},anchor=south}, 
        ]        
	\addplot graphics[xmin=0,ymin=0,xmax=8,ymax=0.02] {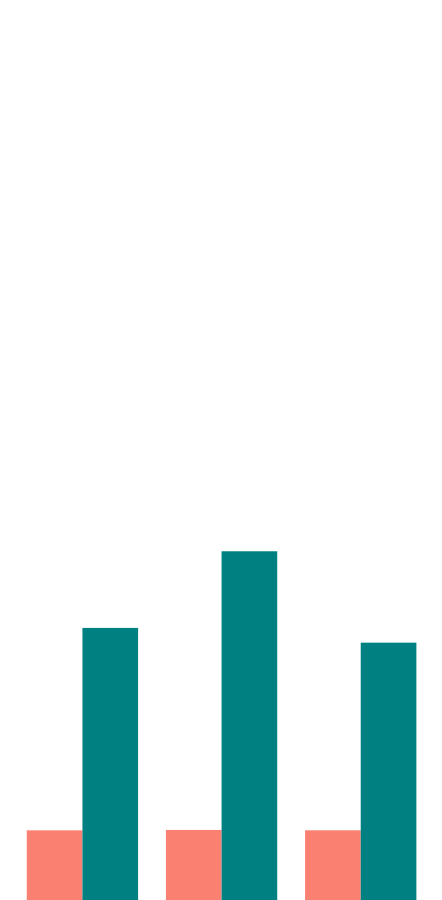};
  \end{axis}
\end{tikzpicture}
   \caption{}
\label{ober_v_error}
\end{subfigure} \hspace{0.05\textwidth}
\begin{subfigure}[t]{0.18\textwidth} 
\begin{tikzpicture}
\begin{axis}[
        axis on top,
        width=\textwidth,
        height = 7em,
        scale only axis,     
        xticklabel style={rotate = 0, align=center, font=\footnotesize, text width = 1.1cm, shift={(0,0 |- {axis description cs:0,0})}},
        ytick={0, 25, 50, 75, 100}, 
        xtick={1.5, 4, 6.5}, 
        xticklabels =  {(i) missing, (ii) \\ erron., (iii) erron. \& missing},        
        ylabel={RMSE [\%]},
        xmin=0,
        xmax=8,
        ymin=0,
        ymax=110,        
        y label style={at={(axis description cs: -0.15,.5)},anchor=south}, 
        ]        
	\addplot graphics[xmin=0,ymin=0,xmax=8,ymax=110] {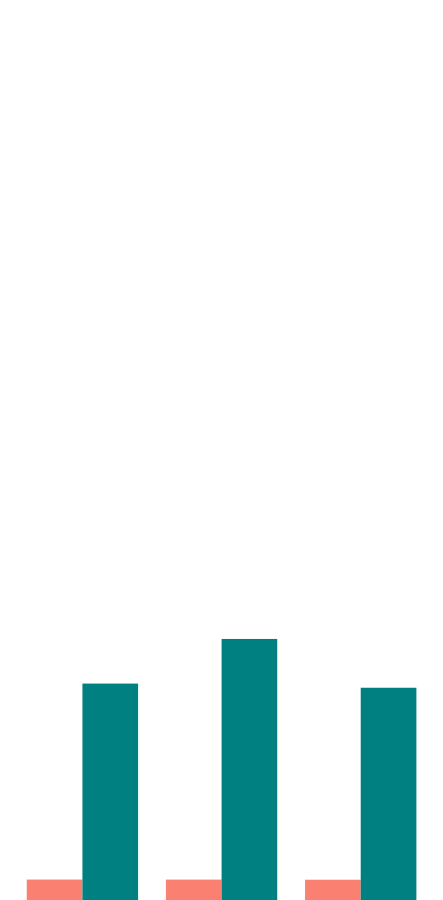};
  \end{axis}
\end{tikzpicture}
   \caption{\centering}
\label{ober_load_error}
\end{subfigure}
\caption{Comparison of RMSE for the estimation of (a) voltage level and (b) line loading between WLS (\protect\includegraphics[height=0.5em]{green.png}) and DSS\textsuperscript{2} (\protect\includegraphics[height=0.5em]{orange.png}) in the $70$-bus network considering (i) missing voltage measurement, (ii) erroneous voltage measurement and erroneous active power flow measurements, and (iii) missing voltage measurement and erroneous voltage measurements.}
\label{ober_rmse_error}
\end{figure}



\begin{figure}
\begin{tikzpicture}
\begin{axis}[
        axis on top,
        width=0.35\textwidth,
        height = 7em,
        scale only axis,
        enlargelimits=false, 
        ytick={1, 1.01, 1.02, 1.03, 1.04},     
        xtick={0,30,60,90,120}, 
        ylabel={Voltage [pu]},
        xlabel={Time [h]},
        xmin=0,
        xmax=120,
        ymin=1,
        ymax=1.04,
        y label style={at={(axis description cs:-0.15,.5)},anchor=south}, 
        ]        
	\addplot graphics[xmin=-5,ymin=0.95,xmax=125,ymax=1.05]    
    {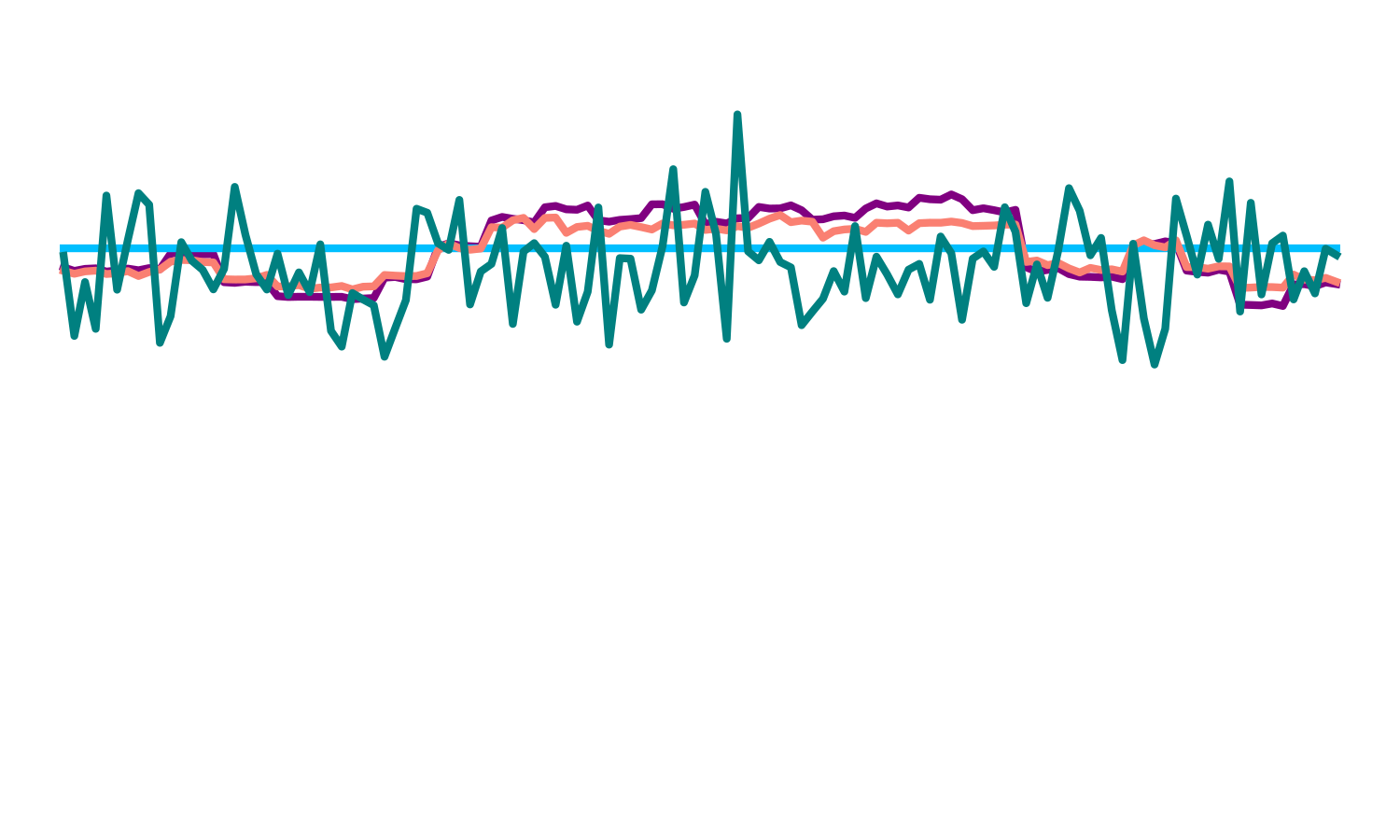};
  \end{axis}
\end{tikzpicture}
\caption{Estimation of voltage level at bus $34$ of the 70-bus network under missing measurement conditions and across the sampling period, using WLS (\protect\includegraphics[height=0.5em]{green.png}), and DSS\textsuperscript{2} (\protect\includegraphics[height=0.5em]{orange.png}). True voltage (\protect\includegraphics[height=0.5em]{purple.png}) and measurement (\protect\includegraphics[height=0.5em]{lightblue.png}) as reference. }
\label{ober_cs7_bus34}
\end{figure}

Fig. \ref{ober_rmse_error} shows the results. The DSS\textsuperscript{2} had high robustness to missing and erroneous measurements in all three cases, with a similar RMSE as the \textit{default} case (no missing or erroneous measurements). However, the erroneous measurement case (ii) impacted the WLS, showing an increase of around $20$\% on relative voltage RMSE. Fig. \ref{ober_cs7_bus34} focuses on one bus with erroneous measurements, the bus $34$ in case (iii). The measurement in bus $34$ was missing for the whole sequence and was naively replaced by the empirical mean value (light blue). 

A key insight of this analysis is that the DSS\textsuperscript{2} was not impacted by this missed value and successfully provided an accurate estimation. Interestingly, the DSS\textsuperscript{2} model was not trained to handle such events. However, using known patterns from neighbouring information, the DSS\textsuperscript{2} remained accurate. Indeed, the GNN architecture increased the interpolation capabilities by incorporating the data symmetries w.r.t. the underlying graph. 


\subsection{Changes in power levels of load and renewables} \label{sec:changes}

\begin{figure}
\begin{subfigure}[t]{0.18\textwidth}
\begin{tikzpicture}
\begin{axis}[
        axis on top,
        width=\textwidth,
        height = 7em,
        scale only axis,
        enlargelimits=false,   
        xticklabel style={rotate = 45, align=right, font=\scriptsize, text width = 1.6cm, shift={(0,0 |- {axis description cs:0.5,-0.1})}, anchor = east},
        ytick={0, 0.005,0.01,0.015,0.02},     
        xtick={1.5, 4, 6.5,9}, 
        xticklabels =  {\textit{default}, ($-30\%,+30\%$), ($+25\%,+100\%$), ($-75\%,+60\%$)},  
        ylabel={RMSE [-]},
        xmin=0,
        xmax=10.5,
        ymin=0,
        ymax=0.02,
        y label style={at={(axis description cs:-0.2,0.5)},anchor=south}, 
        ]        
	\addplot graphics[xmin=0,ymin=0,xmax=10.5,ymax=0.02] {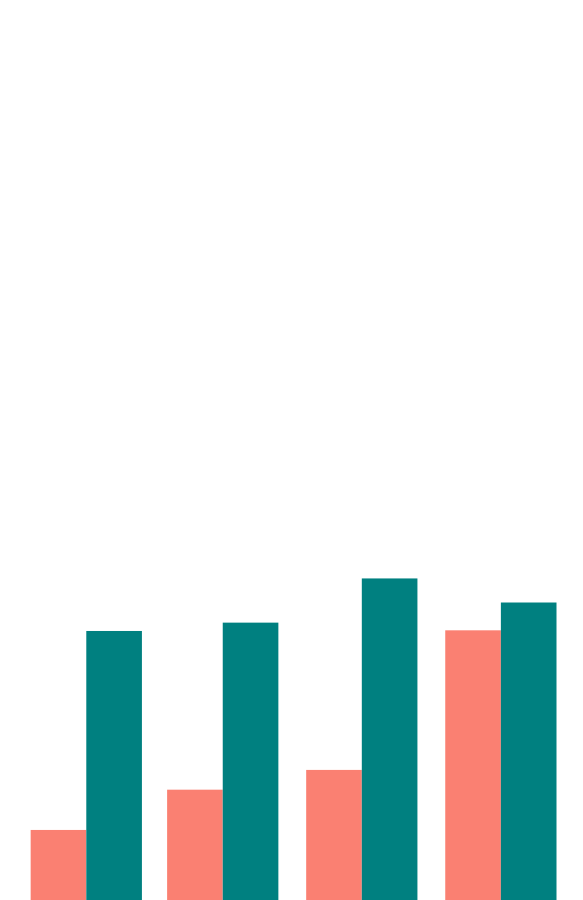};
  \end{axis}
\end{tikzpicture}
   \caption{}
\label{ober_v_disr}
\end{subfigure} \hspace{0.05\textwidth}
\begin{subfigure}[t]{0.18\textwidth} 
\begin{tikzpicture}
\begin{axis}[
        axis on top,
        width=\textwidth,
        height = 7em,
        scale only axis,     
        xticklabel style={rotate = 45, align=right, font=\scriptsize, text width = 1.6cm, shift={(0,0 |- {axis description cs:0.5,-0.1})}, anchor = east},
        ytick={0, 25, 50, 75, 100}, 
        xtick={1.5, 4, 6.5, 9}, 
        xticklabels =  {\textit{default}, ($-30\%,+30\%$), ($+25\%,+100\%$), ($-75\%,+60\%$)},  
        ylabel={RMSE [\%]},
        xmin=0,
        xmax=10.5,
        ymin=0,
        ymax=110,
        y label style={at={(axis description cs: -0.2,0.5)},anchor=south}, 
        ]        
	\addplot graphics[xmin=0,ymin=0,xmax=10.5,ymax=110] {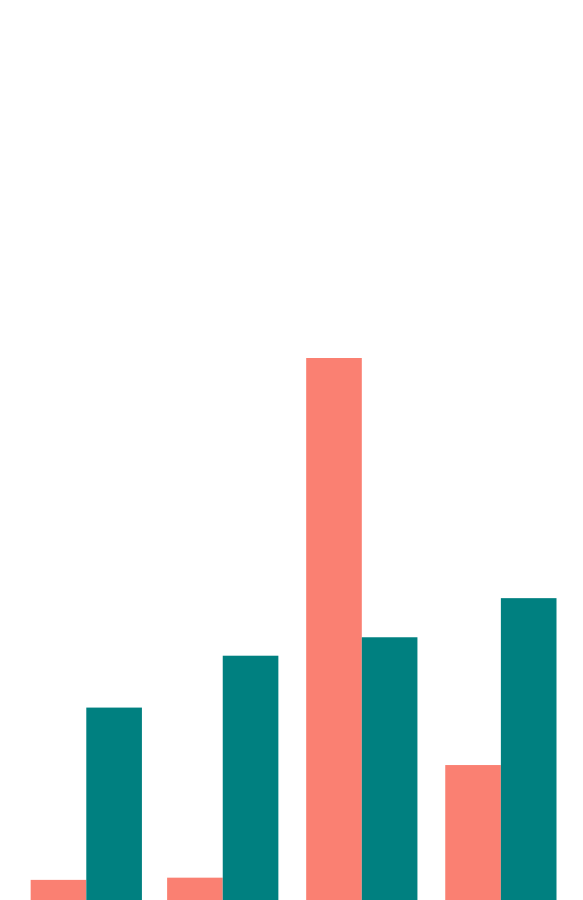};
  \end{axis}
\end{tikzpicture}
   \caption{\centering}
\label{ober_load_dis}
\end{subfigure}
\caption{Comparing RMSE for (a) voltage level and (b) line loading between the WLS (\protect\includegraphics[height=0.5em]{green.png}) and DSS\textsuperscript{2} (\protect\includegraphics[height=0.5em]{orange.png}) in the $70$-bus network considering from the \textit{default} three changes in generation and load: $30$\% decrease in generation and $30$\% increase in load, $25$\% increase in generation and $100$\% increase in load and $75$\% decrease in generation and $60$\% increase in load.}
\label{ober_rmse_dis}
\end{figure}


This case study investigates the generalization capability of the DSS\textsuperscript{2} (and the WLS) to changes in levels of power in the loads and distributed generation compared to the training dataset. Three cases altered the power levels for the testing dataset on the $70$-bus network by:

\begin{enumerate}[leftmargin=2.7cm]
\item[($-30\%,+30\%$)] $30$\% decrease in generation, $30$\% increase in load,
\item[($+25\%,+100\%$)] $25$\% increase in generation, $100$\% increase in load to simulate a system near overload.
\item[($-75\%,+60\%$)] $75$\% decrease in generation, $60$\% increase in load to simulate more voltage deviations
\end{enumerate}
Note that the  DSS\textsuperscript{2} model was never trained on such cases; only \textit{default} power levels were used for training.


Fig. \ref{ober_rmse_dis} shows the results. In the case of a `small' load change ($-30\%,+30\%$), the DSS\textsuperscript{2} showed good estimation performances with only a small increase in RMSE. However, in the cases ($+25\%,+100\%$) and ($-75\%,+60\%$) the RMSE significantly increased. The lines were highly loaded in the case ($+25\%,+100\%$). Hence, the loading estimation was highly impacted. In the case ($-75\%,+60\%$), the deviation in voltage was more harmful to the voltage estimation. These results explored the limitations of the changes in loading levels that the DSS\textsuperscript{2} model could handle. Good results were perceived for changes in loads of around $30$\% showing good generalization capability of the DSS\textsuperscript{2} model to handle state estimation tasks under limited uncertain changes. However, the model became sensitive when the network was extremely loaded or under strong voltage deviations, and then, the model does not generalize well anymore to extreme conditions.

\section{Discussion and conclusions} 
\label{sec:concl}
This paper introduces the Deep Statistical Solver for Distribution System State Estimation. This Deep Learning architecture incorporates the power flow equations in the loss function for physics awareness. Our proposed DSS\textsuperscript{2} approach uses the same objective function as the WLS, allowing to train of the model with a noisy and poorly labelled dataset. This approach is called weak supervision learning, and we combine it with a GNN architecture to enhance the learning from local patterns and the robustness of the model. 
A remarkable advantage of the DSS\textsuperscript{2} is that the larger the power network, the better the performance. The DSS\textsuperscript{2} is based on a GNN architecture that learns from local patterns (in the neighbourhood of buses). Hence, the larger the network, the more local patterns the GNN-based architecture can learn from. We consider this remarkable as conventional power system analysis, for example, for estimating the state, often scales poorly with network size, whereas DSS\textsuperscript{2} showed the reverse effect. Another outstanding advantage is that through learning in the neighbourhood of buses, the DSS\textsuperscript{2} model becomes robust and invariant to changes in individual values, such as missing, erroneous measurements. This is an important practical advantage over other conventional methods (and the studied supervised models) that depend on the accuracy of individual measurements. Our different case studies show that the DSS\textsuperscript{2} is faster, more robust, and more scalable than WLS \textcolor{black}{as DSS\textsuperscript{2} does not involve iterative algorithms and learns from local patterns and noisy measurements}. Compared to supervised models, the weakly supervised DSS\textsuperscript{2} shows equivalent speed and voltage accuracy while outperforming the supervised models in estimating indirect values such as line loading. We conclude that learning from the power flow equations and the neighbourhood are the strengths of DSS\textsuperscript{2}; these incorporate a coupling between voltage magnitudes and voltage angles to fit the measurements. Finally, the DSS\textsuperscript{2} model does not require labels as the approach is weakly supervised learning from the power flow equations. This type of learning makes the DSS\textsuperscript{2} model more practical than other ML-based approaches as labels are scarce.




Our implementation of the DSS\textsuperscript{2} has limitations. First of all, the penalization method used in training impacts the quality of estimation but does not ensure any guarantee of convergence during testing.  \textcolor{black}{Feasible solutions cannot be guaranteed with a data-driven method that focuses on individual accuracy}. Secondly, in our implementation of the H2MG architecture, the assumption to modelling transformers as lines may have particularly limited the accuracy of transformers' loading estimation. There, the model was `forced' to learn a similar input-output mapping for lines and transformers that may reduce the expressivity of the model. Then, the DSS\textsuperscript{2}'s estimation is impacted when the load power level in the network varies significantly. The generalization ability of the DSS\textsuperscript{2} showed a limit of around $30$\% load changes. The changes in measurements are encouraging but should be improved.

Future work could investigate the types of measurements and meter placement decisions that would maximize the DSS\textsuperscript{2} performances. Adding an algorithm that detects changes in the data could benefit quantifying the confidence of state estimations by DSS\textsuperscript{2}. Combining the DSS\textsuperscript{2} for state estimation to a state-of-the-art anomaly detector could improve generalization. \textcolor{black}{Also, an extension to unbalanced systems is deemed possible by extending to unbalanced systems modeling and power flow equations, and it should be investigated in the future}. Then, the network's model in the deep learning architecture could be improved. The proposed model is simple; however, the H2MGNN architecture allows for advanced modelling of components that can further increase the expressivity and performance of the  DSS\textsuperscript{2}. \textcolor{black}{Finally, future work should explore robustness to model inaccuracies and implementation for distribution grids that undergo topology changes. This can be achieved by leveraging the robustness of GNN to graph variation. Such an implementation will further improve the practicality, robustness, and accuracy of the model.}

\bibliographystyle{IEEEtran}
\bibliography{ref.bib}


\end{document}